\documentclass[lettersize,journal]{IEEEtran}
\usepackage{amsmath,amsfonts}
\usepackage{algorithmic}
\usepackage{array}
\usepackage[caption=false,font=normalsize,labelfont=sf,textfont=sf]{subfig}
\usepackage{textcomp}
\usepackage{stfloats}
\usepackage{url}
\usepackage{multirow}
\usepackage{verbatim}
\usepackage{graphicx}
\usepackage{color}
\def\BibTeX{{\rm B\kern-.05em{\sc i\kern-.025em b}\kern-.08em
    T\kern-.1667em\lower.7ex\hbox{E}\kern-.125emX}}
\usepackage{balance}
\begin{document}
\title{Pyramid Pixel Context Adaption Network for Medical Image Classification with Supervised Contrastive Learning
}
\author{Xiaoqing~Zhang,  Zunjie~Xiao, Xiao~Wu, Yanlin~Chen, Jilu~Zhao, Yan~Hu, Jiang~Liu ~\IEEEmembership{Senior~Member,~IEEE}
\thanks{This work was supported in part by National Natural Science Foundation of China (No.82272086), Guangdong Provincial Key Laboratory (No.2020B121201001) and the Stable Support Plan Program (No. 20200925174052004). (Corresponding author:  Jiang~Liu)}

\IEEEcompsocitemizethanks{\IEEEcompsocthanksitem 
Xiaoqing~Zhang and Jiang Liu are with Research Institute of Trustworthy Autonomous Systems, Southern University of Science and Technology, Shenzhen, 518055, China.\protect\\
(e-mail: 11930927@mail.sustech.edu.cn, liuj@sustech.edu.cn)
\IEEEcompsocthanksitem Xiaoqing~Zhang, Zunjie~Xiao, Xiao~Wu, Yanlin~Chen, Yan~Hu, and Jiang Liu were with Department of Computer Science and Engineering, Southern University of Science and Technology, Shenzhen, 518055, China. (e-mail: 11930927@mail.sustech.edu.cn, 
11930387@mail.sustech.edu.cn,11912803@mail.sustech.edu.cn, 12012213@mail.sustech.edu.cn, 11912821@mail.sustech.edu.cn, huy3@sustech.edu.cn, liuj@sustech.edu.cn;
Xiaoqing~Zhang and Zunjie~Xiao contribute equally).
\IEEEcompsocthanksitem  Xiaoqing~Zhang was also with Center for High Performance Computing and Shenzhen Key Laboratory of Intelligent Bioinformatics, Shenzhen Institute of Advanced Technology, Chinese Academy of Sciences, Shenzhen, 518055, China. 
\IEEEcompsocthanksitem Jiang Liu was also with Guangdong Provincial Key Laboratory of Brain-inspired Intelligent Computation, Southern University of Science and Technology, Shenzhen, 518055, China; Singapore Eye Research Institute, 169856, Singapore.}%
}

\markboth{Journal of \LaTeX\ Class Files,~Vol.~18, No.~9, September~2020}%
{How to Use the IEEEtran \LaTeX \ Templates}

\maketitle

\begin{abstract}
Spatial attention mechanism has been widely incorporated into deep neural networks (DNNs), significantly lifting the performance in computer vision tasks via long-range dependency modeling. However, it may perform poorly in medical image analysis. Unfortunately, existing efforts are often unaware that long-range dependency modeling has limitations in highlighting subtle lesion regions. To overcome this limitation, we propose a practical yet lightweight architectural unit, Pyramid Pixel Context Adaption (PPCA) module, which exploits multi-scale pixel context information to recalibrate pixel position in a pixel-independent manner dynamically. PPCA first applies a well-designed cross-channel pyramid pooling to aggregate multi-scale pixel context information, then eliminates the inconsistency among them by the well-designed pixel normalization, and finally estimates per pixel attention weight via a pixel context integration. By embedding PPCA into a DNN with negligible overhead, the PPCANet is developed for medical image classification. In addition, we introduce supervised contrastive learning to enhance feature representation by exploiting the potential of label information via supervised contrastive loss. The extensive experiments on six medical image datasets show that PPCANet outperforms state-of-the-art attention-based networks and recent deep neural networks. We also provide visual analysis and ablation study to explain the behavior of PPCANet in the decision-making process.
\end{abstract}


\section{Introduction}
\label{sec1}
\IEEEPARstart{A}{ttention}  mechanism has achieved remarkable success in a variety of computer vision tasks, \cite{fu2020scene,guo2022attention,Jie2019Squeeze,guo2022beyond,wang2018non}, e.g., object detection, instance segmentation, and image classification. The core idea of the attention mechanism is to guide deep neural networks (DNNs), especially, convolutional neural networks (CNNs), to focus on the informative regions and ignore redundant ones. One of the most representative works is the non-local network (NLNet) \cite{wang2018non}, which belongs to the spatial attention mechanism by explicitly capturing long-range dependencies between pixel positions via a self-attention mechanism. Following the self-attention mechanism, researchers are dedicated to improving self-mechanism design for capturing more sophisticated long-range dependencies among pixel positions \cite{zhu2019asymmetric,fu2019dual,mei2021image,huang2021novel}. Although self-attention-based methods have achieved surpassing performance in a variety of natural image-based tasks, they may not perform well on medical image analysis tasks \cite{rao2021studying}.

\begin{figure}[t]
  \centering
   \includegraphics[width=1.0\linewidth,height=6cm]{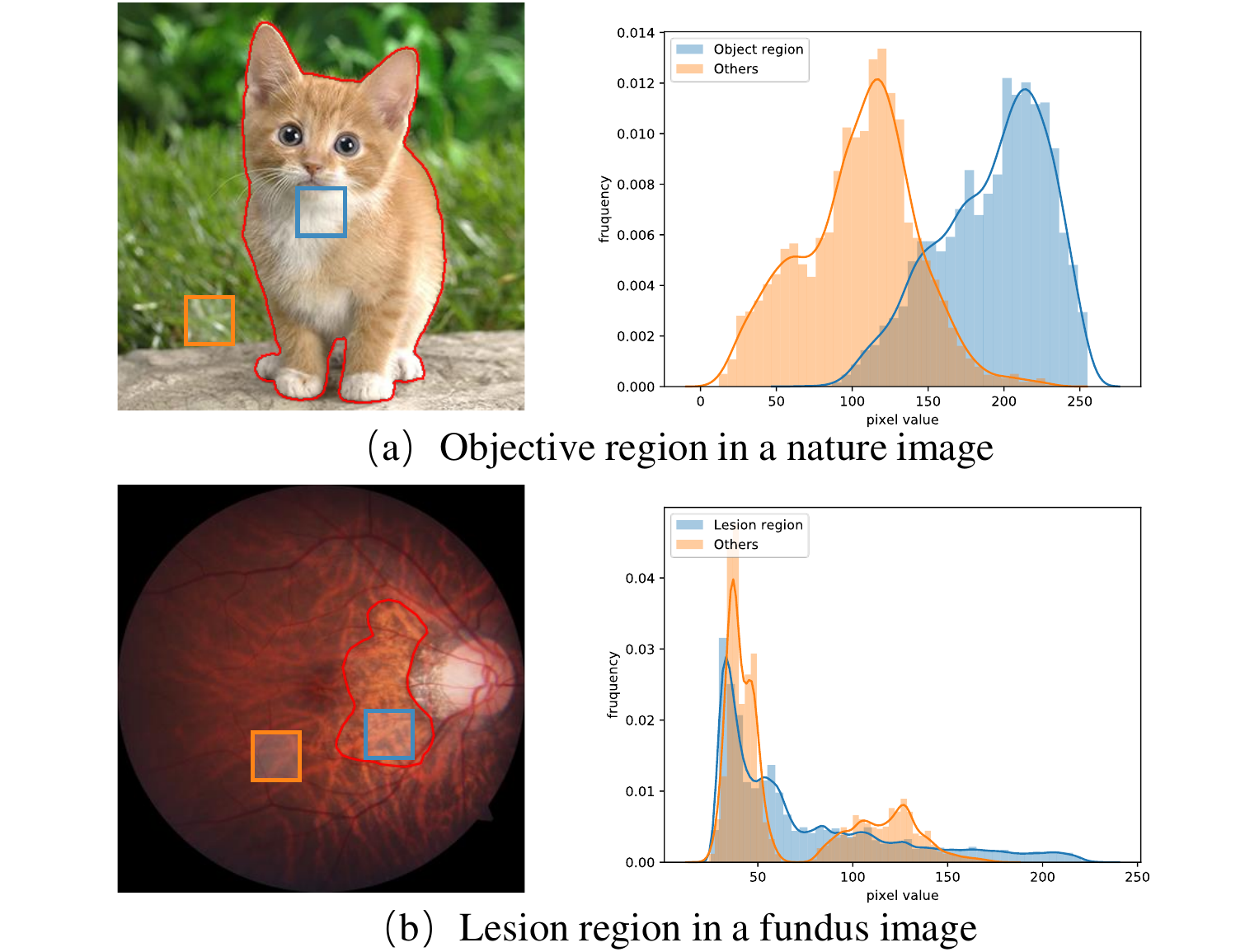}

   \caption{ (a) Object region in a natural image, which is salient through observing pixel value distribution difference between object region and other regions. (b) The subtle lesion region of myopia on the fundus image. We also present a pixel value distribution comparison between a subtle lesion region and a redundant region.}
   \label{fig:1}
\end{figure}

\begin{figure}[t]
  \centering
   \includegraphics[width=1.0\linewidth]{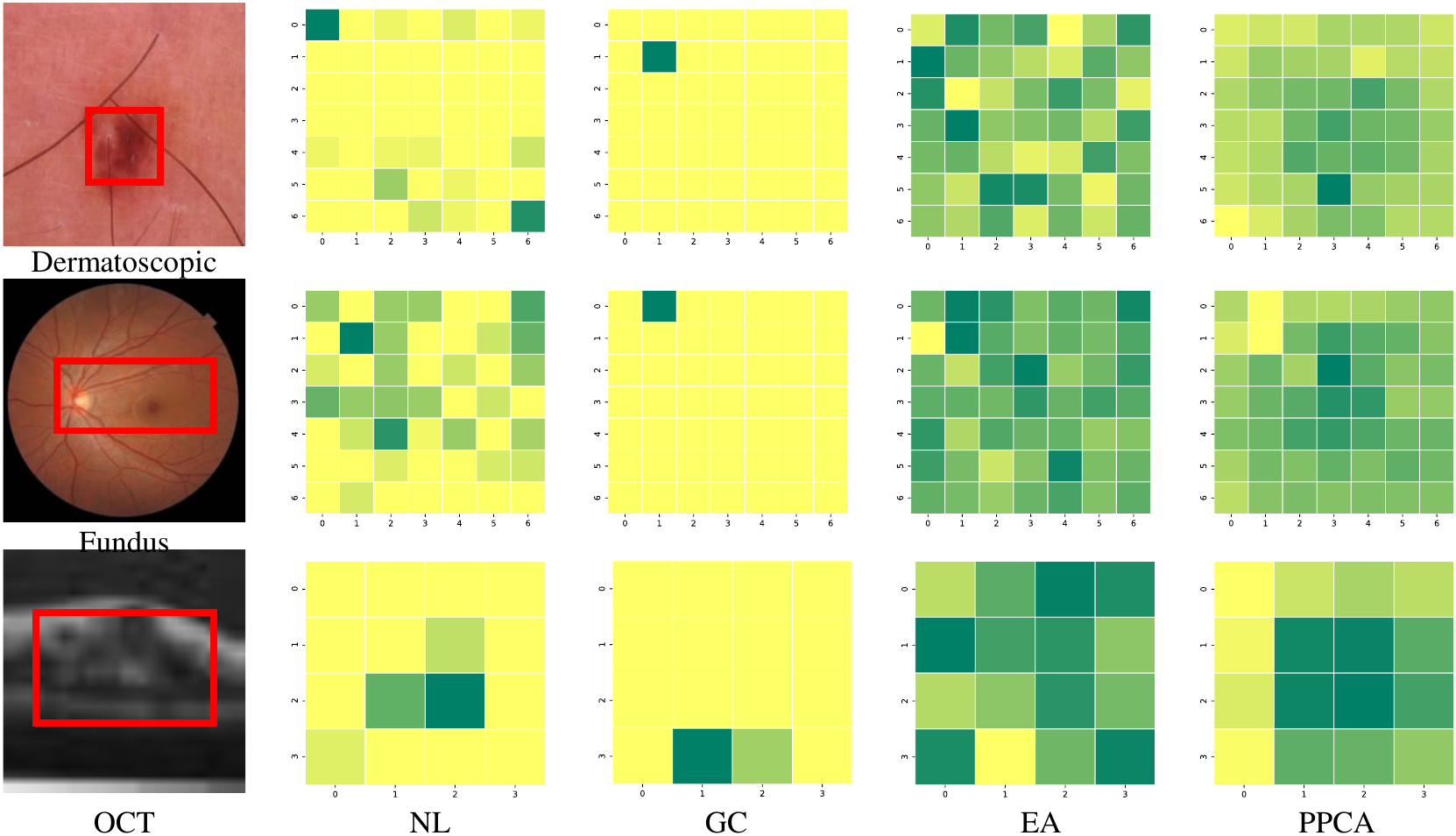}

   \caption{Pixel attention weight maps generated by NL \cite{wang2018non}, GC \cite{cao2019gcnet}, EA \cite{guo2022beyond}, and PPCA at the high stage of ResNet18 for skin disease, blinding disease, and retinal disease based on three medical image modalities: dermatoscopic image, fundus image, and optical coherence tomography (OCT) image. Clearly, our method is more capable of emphasizing subtle lesion regions accurately (red box) than state-of-the-art spatial attention methods. }
   \label{fig:1-1}
\end{figure}

In seeking answers to this phenomenon, we have gained insights as follows: (1) \textbf{Long-Range Dependency Modeling.} The Self-attention mechanism usually captures long-range dependencies across all pixel positions to learn such a pixel position correlation, inevitably introducing redundant position information from other pixel positions. The negative influences of redundant position information on natural image-based learning tasks can be ignored. This is mainly because the object region in natural images is salient (as shown in Fig.~\ref{fig:1}(a)), which is easily highlighted by long-range dependency capturing. In contrast, lesion regions in medical images are subtle. That is, pixel context information difference between redundant regions and lesion regions is obscure \cite{haghighi2021transferable, haghighi2022dira}  (as shown in Fig.~ \ref{fig:1}(b)), which makes it difficult to emphasize lesion region information through modeling long-range dependency. In particular, Fig.~\ref{fig:1}(b) provides a pixel value distribution comparison between the subtle lesion region of myopia and the redundant region on the fundus image along the spatial dimension; we see the pixel value distributions between these regions are similar, keeping consistent with the above analysis.
(2) \textbf{Pixel Context Aggregation.} Channel attention methods have aggregated multi-scale spatial context information to improve performance with spatial pyramid pooling method \cite{hu2020span,li2018pyramid,guo2020spanet,wang2021pyramid,wang2021pvtv2}. However, existing spatial attention methods only have utilized pointwise convolution (Conv$1 \times 1$) \cite{wang2018non}, or individual cross-channel pooling (CP) \cite{chen2019condensation} methods to aggregate single-scale pixel context information along the channel axis, inevitably ignoring the significance of multi-scale pixel context information aggregation. According to our extensive literature survey, we have found that no spatial attention method has exploited the potential of multi-scale pixel context information to improve representational ability of DNNs.

Based on the above systematical analysis, this paper is really curious to find out: \textit{(1) Can one learn an alternative method to highlight informative pixel positions and suppress trivial ones without capturing long-range dependencies among pixel positions in the spatial attention module? (2) Can we incorporate \textbf{multi-scale pixel context} information into spatial attention design to improve the performance and explanation of CNNs?}

To answer these two questions, we propose a novel yet lightweight architectural unit, Pyramid Pixel Context Adaption (PPCA) module, which explicitly incorporates multi-scale pixel context information into CNN representations through a form of pixel-independent context recalibration. Our PPCA consists of a triplet of components: \textit{Cross-Channel Pyramid Pooling}, \textit{Pixel Normalization}, and \textit{Pixel Context Adaption}. To the best of our knowledge, this paper is the first to design a \textit{Cross-Channel Pyramid Pooling} for aggregating multi-scale pixel context information at the same pixel positions through different cross-channel scales at the channel dimension. Note that multi-scale pixel context features are extracted at each pixel position, and only specific-scale pixel context plays a significant role. Then, \textit{Pixel Normalization} is developed to eliminate the significant fluctuation of multi-scale pixel context distribution per pixel position, which is different from previous normalization methods that perform the pixel context statistics at the feature maps. It is followed by \textit{Pixel Context Adaption}, which adaptively fuses normalized multi-scale pixel context information to produce pixel attention weights via pixel-level operation. The pixel attention weights are finally supposed to reweigh per pixel position to emphasize or ignore their information. 
Our PPCA only increases negligible computational cost and few parameters, and we plug PPCA module into modern CNNs, constructing a novel network architecture for medical image classification, named PPCANet. Furthermore, to boost the medical classification performance of PPCANet, we apply supervised contrastive learning \cite{khosla2020supervised} to exploit the label information through the contrastive pairs.

To demonstrate the effectiveness and efficiency of our method, we conduct extensive experiments on six medical image datasets. The results show that our method is superior to state-of-the-art (SOTA) attention methods and recent deep neural networks. Beyond the practical improvements, we empirically analyze the effects of the pixel level adaption on emphasizing significant pixel positions and redundant ones through visual analysis and ablation study: it controls the relative contributions of multi-scale pixel context information and pixel normalization, which is beneficial to improve the interpretability of DNNs in the decision-making process. Fig.~\ref{fig:1-1} provides the generated pixel attention weight maps of PPCA and other SOTA attention methods, showing that our method can more accurately locate subtle lesion regions (red box) than others, agreeing with the clinician’s diagnosis process.

In summary, the main contributions of this paper are as follows:
\begin{itemize}
\item We propose a pyramid pixel context adaption (PPCA) module to improve the representational capability of CNNs by combining multi-scale pixel context information and pixel normalization method. In particular, this paper is the first to develop a cross-channel pyramid pooling method to aggregate and exploit multi-scale pixel context information to boost performance via the spatial attention method. Additionally, we design a pixel normalization method to eliminate the inconsistency of multi-scale pixel context information per pixel position.
\item We combine the PPCA module with modern CNNs to construct the PPCANet for medical image classification. PPCANet can highlight subtle lesion regions efficiently, which is beneficial for improving medical image classification. In addition, we utilize supervised contrastive learning to exploit the label information to further enhance performance from the contrastive pair perspective.
\item The comprehensive experiments on six medical image classification tasks consistently demonstrate the superiority and generalization capability over SOTA attention methods. Moreover, visual analysis and ablation study are implemented to interpret the inherent decision-making behavior of our PPCA, conducing to enhancing the explanation in the decision-making process.
\end{itemize}

The rest of this paper is organized as follows. Section~\ref{sec:2} briefly reviews pyramid pooling, normalization, and attention mechanism. Section~\ref{sec:3} introduces our proposed method in detail. Section~\ref{sec:4} presents the dataset description, experiment settings, results and analysis. We discuss and conclude our method in Section~\ref{sec:5} and ~\ref{sec:6}.

\section{Related Work}
\label{sec:2}

\subsection{Pyramid Pooling} 
Pyramid pooling is a widely acknowledged technique to extract multi-scale context information \cite{wu2022p2t,lian2021cascaded,li2018pyramid,zhu2019asymmetric}. Currently, spatial pyramid pooling has been widely utilized in various tasks, e.g., image classification, semantic segmentation, and object detection. He et al. \cite{he2015spatial} present spatial pyramid pooling to obtain multi-scale spatial context information for image classification. Gu et al. \cite{gu2019net} propose spatial pyramid pooling for semantic segmentation. Guo et al. \cite{guo2020spanet} propose a spatial pyramid attention (SPA) module by incorporating spatial pyramid pooling for image classification. Wang et al.~\cite{wang2021pyramid,wang2021pvtv2} embed spatial pyramid pooling into the vision transformer (ViT) for dense prediction. 
Unlike existing works that apply spatial pyramid pooling to extract multi-scale spatial context information for channel attention and ViT architecture design, we propose a cross-channel pyramid pooling to extract multi-scale pixel context information for spatial attention design, which has not been studied before.

\begin{figure*}[t]
	\centering
	\centerline{\includegraphics[width=0.92\linewidth]{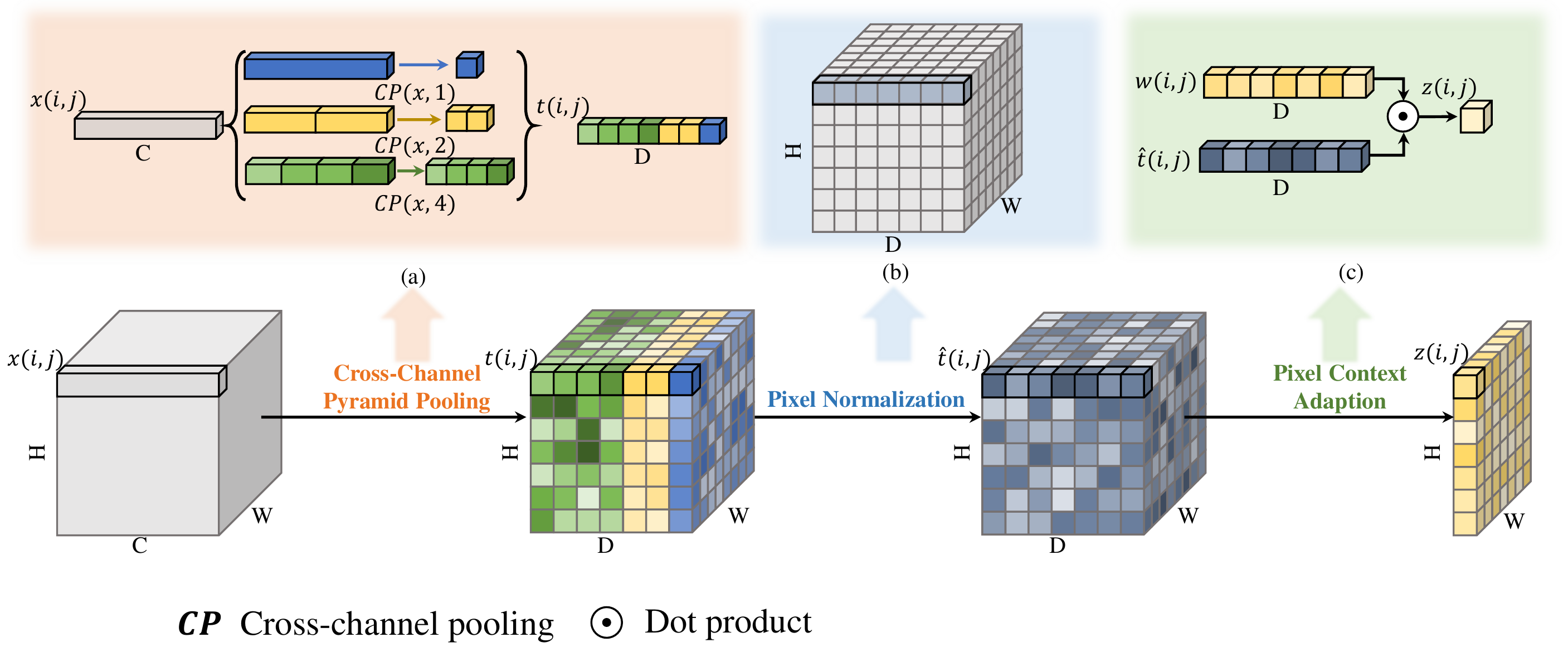}}
	\caption{The detailed construction of pyramid pixel context recalibration (PPCA) module.  Given the intermediate feature maps $X \in R^{C \times H \times W}$, PPCA generates the pixel attention weight map $G\in R^{1 \times H  \times W}$.}
	\label{fig:2}
\end{figure*}

\subsection{Normalization} 
Batch normalization (BN) \cite{ioffe2015batch} is a pioneered technique that normalizes the statistics along the batch axis to stabilize the intermediate feature distribution of hidden layers, allowing deep neural networks to train faster yet fluctuate smaller. Additionally, the property of batch size in BN dramatically affects the network performance when reducing batch size due to inaccurate batch statistics estimation. Several normalization methods have been proposed to tackle this issue \cite{ba2016layer, wen2019exploiting,wu2018group,ortiz2020local,gao2021representative}. Layer normalization (LN) \cite{ba2016layer} computes the statistics along the channel axis, and instance normalization (IN) \cite{ulyanov2016instance} performs the BN-like normalization operator per intermediate feature map. Weight normalization (WN) \cite{salimans2016weight} normalizes the filter weights. Group normalization (GN) \cite{ wu2018group} divides feature maps into several groups and then computes the statistics per group. Positional normalization (PONO) \cite{li2019positional} normalizes the per-pixel position independently across the channels by mining the first and second moments of statistics information. 

The design of our PPCA is motivated by LN and PONO, we propose a pixel normalization operator to normalize the first order moment pixel context statistics for each along the channel axis at the same pixel positions, aiming to emphasize significant pixel positions and suppress trivial ones in an efficient manner.

\subsection{Attention Mechanism} 

The current research directions of attention mechanism can be roughly divided into three categories \cite{fu2020scene,guo2022attention,li2019selective, fu2020scene, zhang2022attention, huang2019ccnet,schlemper2019attention,zheng2021rethinking}: channel attention, spatial attention, and combination. Squeeze-and-excitation (SE) \cite{Jie2019Squeeze} is one of the successful channel attentions, which captures long-range dependencies among channels. 
Considering PPCA belongs to the spatial attention mechanism, this paper briefly surveyed spatial attention modules. Coordinate attention (CA) \cite{hou2021coordinate} learns long-range spatial context information embedding positional information into channel attention. Convolutional block attention module (CBAM) \cite{woo2018cbam} utilizes a spatial attention (SA) block to capture local-range spatial context information. Recently, self-attention mechanism and its variants  \cite{bello2019attention,cao2019gcnet,guo2022beyond,zhao2018psanet,chu2021twins} dominates the spatial attention research due to their powerful capability in modeling long-range dependencies among all pixel positions. For instance, the global context network (GCNet) \cite{cao2019gcnet} utilizes a self-attention mechanism to construct a global context (GC) block. Bello et al. \cite{Bello_2019_ICCV} uses multihead attention (MHA) for image classification. External attention (EA) \cite{guo2022beyond} applies the learnable weights to each pixel context and constructs the pixel context long-range dependency via a self-attention mechanism. However, most existing spatial attention methods capture long-range dependencies among all pixel positions, which is skilled at highlighting concentrative object regions in natural images and may be poor at learning subtle lesion regions in medical images.

In contrast to these methods dedicated to designing self-attention-based spatial attention modules, our method designs a more efficient yet lightweight way to highlight or suppress pixel positions in a pixel-independent manner by incorporating multi-scale pixel context information.

\section{Methodology}
\label{sec:3}

\subsection{Pyramid Pixel Context Adaption Module}
\label{sec:method}
Given the intermediate feature maps $X \in R^{C \times H \times W}$, PPCA generates the pixel attention weight map $G\in R^{1 \times H  \times W}$, where C, H and W indicate the number of channels, height and width of feature maps accordingly. As illustrated in Fig.~\ref{fig:2}, 
our method is abstracted by the following three components: \textit{Cross-Channel Pyramid Pooling}, \textit{Pixel Normalization}, and \textit{Pixel Context Integration}.

\subsubsection{Cross-Channel Pyramid Pooling}
\label{sec:cpp}
Spatial pyramid pooling is often used in channel attention methods to aggregate multi-scale spatial context information from multi-scale feature map regions, significantly improving performance. However, current spatial attention methods only aggregate single-scale pixel context information with cross-channel pooling and have yet to exploit the potential of multi-scale pixel context information. To address this problem, we propose a cross-channel pyramid pooling (CCPP) method to aggregate multi-scale pixel context information of all pixel positions from different cross-channel scales. Sophisticated CCPP design can be used to further boost performance, but this is not the fundamental goal of this paper. Thus, we simply employ an averaged CCPP to aggregate the multi-scale pixel context features of all pixel positions from three different cross-channel scales along the channel dimension. Fig.~\ref{fig:2}(a) provides a visual implementation case of CCPP with three cross-channel scales at the pixel position $x(i,j)$, which can help audiences understand the proposed CCPP easily. The output of multi-scale pixel context description $T \in R^{D \times H \times W}$ ($D$ is the number of pixel context feature maps, and $D$ is equal to 7 in this paper based on the experimental results) of all pixel positions through the CCPP operator can be computed by:
\begin{equation}
\begin{aligned}
  T = [CP(X, 1), CP(X, 2), CP(X, 4)],
  \label{eq:1}
 \end{aligned}
\end{equation}
where $CP(X, 1) $, $CP(X, 2)$, and $CP(X, 4)$ indicate one , two, and four pixel context feature maps extracted from three different cross-channel scales. 
$CP$ indicates the cross-channel pooling, which performs at per pixel position $x(i,j)$ across channels $K \leq C$  can be computed as follows:
\begin{equation}
 \mu(i,j)= \frac{1}{K}\sum_{k=1}^{K}x(k,i,j), 
\label{eq:2}
\end{equation}
where $\mu(i,j)$ is averaged pixel context feature of $x(i,j)$. Furthermore, the number of cross-channel scales in CCPP can be extended to other cases according to learning tasks. In the ablation study, we also conduct corresponding ablation experiments.

\subsubsection{Pixel Normalization} 
\label{sec:pixel}
Existing normalization methods such as LN and BN compute the statistics of pixel context features across the feature map, which can not effectively eliminate the inconsistency of multi-scale pixel context features $T$ at the same pixel positions. To stabilize multi-scale pixel context feature distribution, our PPCA introduces a pixel normalization (PN) operator to normalize them across all pixel context feature maps at each pixel position, as illustrated in Fig.~\ref{fig:2}(b). For each pixel context $t_{(d,i,j)}$ , the PN can be formulated as follows:
\begin{equation}
 \hat{t}_{d,i,j} =  \frac{t_{(d,i,j)}-\mu_{(i,j)}^{(t)}}{\delta_{(i,j)}^{(t)}}, 
 \label{eq:3}
\end{equation}
where $\hat{t}_{(d,i,j)}$ indicates the normalized multi-scale pixel context at the pixel position $(i,j)$ of $d$ pixel context feature map. $\mu_{(i,j)}^{(z)}$ and $\delta_{(i,j)}^{(z)}$ are the mean and standard deviation of multi-scale pixel context features at pixel position $t(i,j)$, which can be computed as:
\begin{equation}
\mu_{(i,j)}^{z} = \frac{1}{D}\sum_{d=1}^{D}t_{(d,i,j)}, \delta_{(i,j)}^{(z)} = \sqrt{\frac{1}{D}\sum_{i=1}^{D}(t_{(d,i,j)}-\mu_{(i,j)}^{(t)})^{2}} +\xi,
\label{eq:4}
\end{equation}
where $\xi$ is a very small constant. According to Eqs.~\ref{eq:3}-\ref{eq:4}, we can obtain normalized multi-scale pixel context feature maps $ \hat T \in R^{D \times H \times W}$ based on the PN operation.

\subsubsection{Pixel Context Adaption}
Following the PN, we define the pixel context adaption (PCA) function (as shown in Fig.~\ref{fig:2}(c)) to convert the normalized multi-scale pixel context feature maps $ \hat T \in R^{D \times H \times W}$ into pixel attention weights $G$ via a pixel-wise fully-connected (PFC) layer,  which can be represented by:
\begin{equation}
 Z = W \cdot \hat T, G=\sigma(Z),
\label{eq:5}
\end{equation}
where  $W \in R^{7 \times H \times W} $ indicates learnable parameters; $\sigma$ is the sigmoid function as the gating mechanism; $Z \in R^{1 \times H \times W} $ indicates the encoded multi-scale pixel context features. In the ablation study, we will test the effects of other PCA implementations like Conv $1\times1$, summation, and Conv $5\times5$.

Finally, the augmented feature maps $Y \in R^{ C \times H \times W}$ are computed as follows:
\begin{equation}
Y= G \cdot X. 
\label{eq:7}
\end{equation}

\subsubsection{Complexity Analysis}
Our PPCA is supposed to be lightweight in terms of computational cost and parameters. The PCA function determines the additional parameters of PPCA: $\sum_{s=1}^{S}H_{s}\cdot W_{s} \cdot N_{s} \cdot 7$. $S$ and $N_{s}$ represent the number of stages and the number of repeated blocks in the s-th stage, which we follow the same definition of stage in \cite{he2016deep}, $H_{s}$ and $W_{s}$ represent the height and width of feature maps in the s-th stage. Therefore, the total number of additional parameters for PPCA is:
\begin{equation}
   7\sum_{s=1}^{S} N_{s} \cdot H_{s}\cdot W_{s},
   \label{eq:8}
\end{equation}
which is far less than the total number of parameters for NL:$\frac{2+r}{r}\sum_{s=1}^{S}N_{s}C_{s}^{2}$ where $r$
and $C_{s}$ represent the reduction ratio and the number of output channels in the s-th stage. 
According to Eq.~\ref{eq:8}, the extra parameters of PPCA are determined by the height and width of a feature map, which is different from the extra parameters of channel attention methods determined by the number of channels, e.g., the number of parameters in SE is $\frac{2}{r}\sum_{s=1}^{S}N_{s}C_{s}^{2}$. Theoretically, our method has parameter advantages over channel attention methods on low-resolution images, which will be verified in experiments. As for computational cost, our PPCA introduces negligible extra computational cost compared to original network architectures, which can be computed as follows:
\begin{equation}
   8\sum_{s=1}^{S} N_{s} \cdot H_{s} \cdot W_{s},
   \label{eq:9-1}
\end{equation}
which is far smaller than computational cost of SE ($\frac{4}{r}\sum_{s=1}^{S}N_{s}C_{s}^{2}$) and NL ($4\sum_{s=1}^{S} N_{s} \cdot (W_{s}H_{s})^{2} \cdot C_{s}$).
For example, given a $224 \times 224 $ pixel image or $28 \times 28$ pixel image as input, PPCA-ResNet50 shares almost the same computational cost as ResNet50.

\subsection{Network Architecture}
\begin{figure*}
\centering
       \centerline{\includegraphics[width=0.6\linewidth,height=7cm]{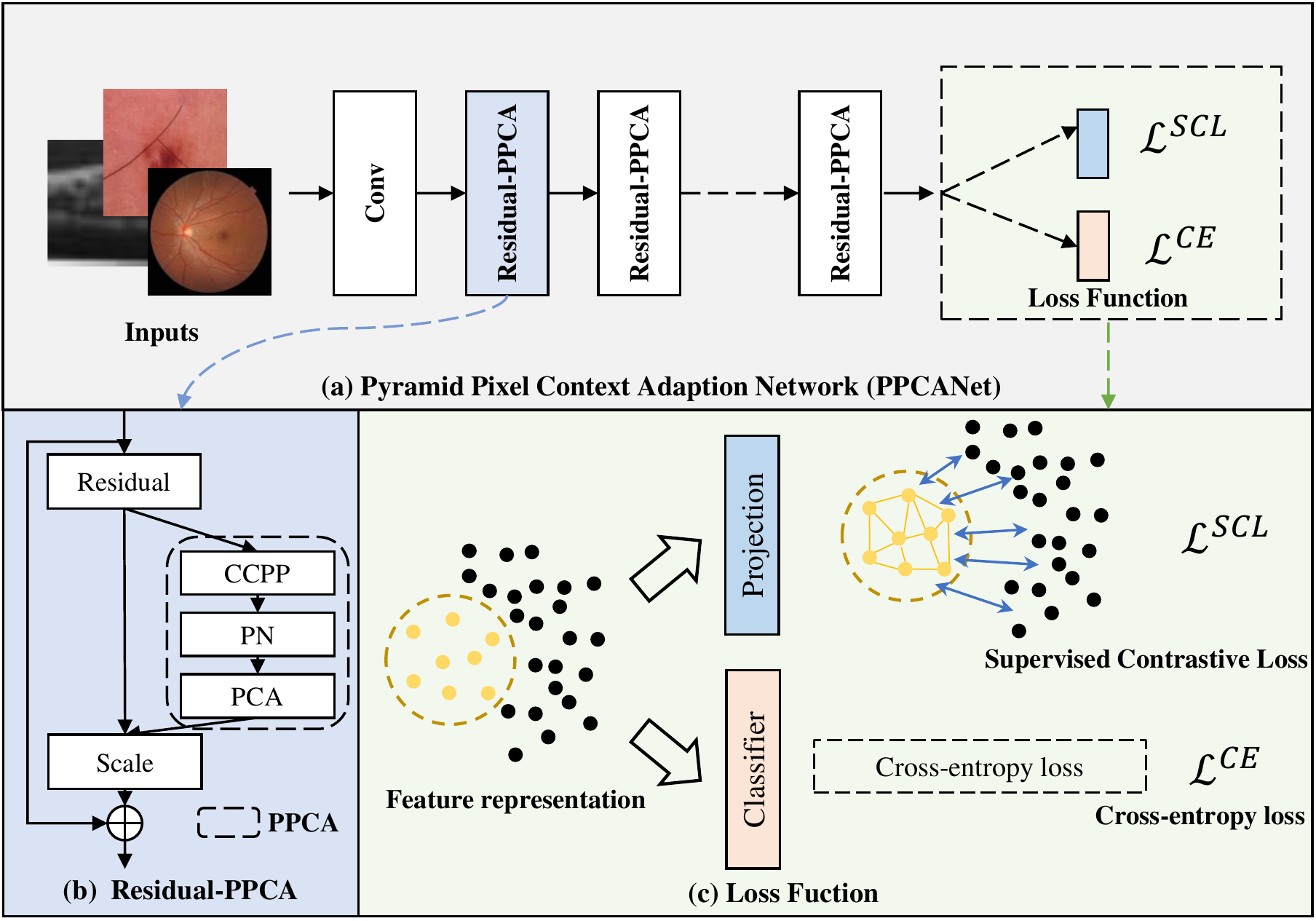}}
  \caption{Pyramid pixel context adaption network (PPCANet) for medical image classification (a), in which we combine the PPCA module with the residual module to construct a Residual-PPCA module (b). Furthermore, to achieve better performance, we adopt supervised contrastive loss as the supplement for cross-entropy loss for further exploiting label information (c).}
	\label{fig:3}
\end{figure*}

Fig.~\ref{fig:3}(a) provides the general framework of pyramid pixel context adaption network (PPCANet) for medical image classification by taking ResNet as the backbone. In the PPCANet, we combine the PPCA module with the residual module to construct a Residual-PPCA module. Our PPCANet takes a medical image as the input, and a convolutional layer is applied to produce low-level feature representations. Next, a sequence of Residual-PPCA modules generate high-level feature representations. Finally, a global average pooling (GAP) and the softmax function are utilized to generate the predicting label of each input medical image. Furthermore, we not only adopt the classical cross-entropy (CE) loss as the loss function but also take supervised contrastive loss as the auxiliary loss to exploit label information from the contrastive pair perspective (as presented in Fig.~\ref{fig:3}(c)), which will be introduced in the following section.

\subsection{Supervised Contrastive Loss}

Recently, contrastive learning has attracted much attention in the self-supervised learning (SSL) field. Its fundamental idea is contrastive loss (CL) by constructing positive sample and negative sample pairs. In SSL, CL can be viewed as another form of triplet loss, which aims to reduce the distance among the feature representations in the positive sample pairs and increase the distance among the feature representations in the negative sample pairs. However, self-supervised CL is unable to leverage label information. To address this problem, this paper attempts to supervise contrastive loss (SCL) to exploit label information to improve the feature representation ability of DNNs from the contrastive pair view, as presented in Fig.~\ref{fig:3}(c). Given the ground truth $Y =\{y_{1}, y_{2},…, y_{N}\}$ and their corresponding samples $X ={x_{1}, x_{2},…, x_{N}}$. The SCL can be computed as follows:
\begin{equation}
   \mathcal{L}_{i}^{\text {SCL}}=-\frac{1}{N_{y_{i}}} \log \frac{\sum_{j=1}^{N} \mathbb{1}_{\left[y_{i}=y_{j}\right]} \exp \left(z_{i} \cdot z_{j} / \tau\right)}{\sum_{k=1}^{N} \mathbb{1}_{[k \neq i]} \exp \left(z_{i} \cdot z_{k} / \tau\right)},
   \label{eq:9}
\end{equation}
where $N_{y_{i}}$ denotes the number of samples that labeled as $y_{i}$, $\mathbb{1} \in {0,1}$ equals to 1 if $y_{i} =y_{j}$, $z_{i}$ 
is the normalized feature representation $x_{i}$, $\tau \in \mathcal{R}^{+}$ is a scalar temperature parameter.

Moreover, CE treats every sample equally, which exploits the label information from the sample-wise perspective. Hence, we propose a hybrid loss (HL) to update the parameters of DNNs by combining CE with SCL for obtaining better performance from both sample-wise and contrastive pair perspectives, which is formulated as follows:
\begin{equation}
   \mathcal{L}_{i}=  \lambda \mathcal{L}_{i}^{\text {CE}} + (1- \lambda ) \mathcal{L}_{i}^{\text {SCL}} ,
   \label{eq:10}
\end{equation}
where $\lambda$ is a hyperparameter, which is determined by the experimental results.

\section{Experiments and Result Analysis}
\label{sec:4}
In this section, we first introduce datasets and implementation details, then demonstrate the effectiveness and generalization ability of our proposed method through comparisons to SOTA attention methods and recent deep neural network architectures. Then, we conduct systematic visual analyses to investigate the inherent behavior of PPCA.

\subsection{Datasets}
Table~\ref{tab:1-1} lists details of six medical image datasets.

\begin{table}
\caption{The details of long-tailed medical datasets.}
\label{tab:1-1}
\centering
\begin{tabular}{c|c|c}
\hline
Datasets& Number of classes&Number of samples \\
\hline
ISIC2018&7& 10,015 \\
iMED-MD&3&1,959 \\
Fundus-Isee &4&10,00\\
OCTMNIST&4&109,309\\
RetinaMNIST&5&1,600\\
BreastMNIST&2&780\\
\hline
\end{tabular}
\end{table}

\textbf{ISIC2018~\cite{tschandl2018ham10000} .} It is a publicly available skin lesion dataset with 10,015 images of seven different labels. This paper uses the same data preprocessing strategy and dataset splitting as in literature \cite{liu2022acpl}.

\textbf{iMED-MD.} It is a fundus image dataset with 1,959 images, consisting of normal, myopia, and diabetic retinopathy (DR). We split it into training and testing subsets. Each image is labeled with normal, myopia, and DR by three experienced ophthalmologists, respectively. The training subset has 1,370 images (700 images of normal, 509 images of myopia, and 161 images of DR), and the testing subset has 589 images (300 images of normal, 219 images of myopia, and 70 images of DR).

\textbf{Fundus-Isee.} It is a fundus image dataset with 10,000 images that contains four different ocular diseases: age-related macular degeneration (AMD) (720), DR (270), glaucoma (450), myopia (790), and normal (7,770). We follow the same data augmentation strategy and dataset splitting method in literature \cite{FANG2021101981}.

\textbf{MedMNIST~\cite{yang2021medmnist}.} MedMNIST is an MNIST-like benchmark for medical image classification, containing 15 medical image datasets. In this paper, we use three MedMNIST datasets to further demonstrate the efficiency and effectiveness of our PPCA: OCTMNIST, RetinaMNIST, and BreastMNIST. OCTMNIST comprises 109,309 OCT images of four retinal diseases. RetinaMNIST contains 1,600 retina fundus images of five DR severity levels. BreastMNIST has 780 breast ultrasound images of two labels. The image size of these three datasets is $28 \times 28$. Moreover, data augmentation and dataset splitting methods are adopted from literature \cite{yang2021medmnist} for a fair comparison.

\subsection{Implementation details}
\label{sec:exper}

\textbf{Baselines:}
In this paper, we use the following SOTA attention methods to demonstrate the effectiveness of PPCA based on six medical datasets, including SE, SPA, CA, SA (used in CBAM), NL, EA, and GC, by adopting two commonly used CNN architectures as backbones: ResNet18 and ResNet50 \cite{he2016deep}. Specifically, \textit{SA, NL, EA, and GC} are spatial attention methods involving local-range and long-range dependency modeling, which are able to verify the superiority of our method comprehensively. Furthermore, we also adopt recent deep neural networks including specifically designed for medical image classification, SSFormer~\cite{wang2022stepwise}, CABNet~\cite{he2020cabnet}, RIRNet ~\cite{ZHANG2022102499}, RCRNet~\cite{zhang2024regional}, DANet~\cite{fu2020scene},
vision transformer (ViT) \cite{dosovitskiy2020image}, Res-MLP ~\cite{touvron2021resmlp}, swin transformer (Swin-T) ~\cite{liu2021swin}, CoAtNet~\cite{dai2021coatnet}, MLP-Mixer~\cite{tolstikhin2021mlp}, AANet~\cite{Bello_2019_ICCV}, ConvNeXt~\cite{liu2022convnet}, MetaFormer~\cite{yu2022metaformer}, and pyramid vision transformer (PVT) ~\cite{wang2021pvtv2} for comparison.

\textbf{Experiment setup:}
These methods are implemented by the PyTorch package and use SGD optimizer with default settings during the training process. The initial learning rate is set to 0.025 and is decreased by a factor of 5 per 25 epochs. We set batch size and epochs to 32 and 150 accordingly and run all methods on two TITAN V NVIDIA GPUs under the same experiment settings. The code of this paper will be released soon.

\textbf{Evaluation metrics:}

Five commonly accepted evaluation metrics are adopted to evaluate the performance and model complexity of PPCA, SOTA attention methods, and baselines: accuracy (ACC), the area under the ROC curve (AUC), F1, parameters (Params.), and GFLOPs. Particularly, compared to advanced DNNs, we only adopt ACC, AUC, and F1 to assess their performance.

\begin{table*}[t]
  \caption{Performance comparison of different attention methods on three medical image datasets (ISIC2018, iMED-MD, and Fundus-Isee) in terms of  accuracy, AUC, F1, parameters, and GFLOPs.}
  \centering
  \begin{tabular}{cccc|ccc|ccc|cc}
    \hline
    \multirow{2}{*}{Method}& \multicolumn{3}{c}{ISIC2018}&  \multicolumn{3}{c}{iMED-MD}&  \multicolumn{3}{c}{Fundus-Isee}&\multirow{2}{*}{Params}& \multirow{2}{*}{GFLOPs}\\
     & ACC&AUC&F1&ACC&AUC&F1&ACC&AUC&F1\\
    \hline
    ResNet18 &78.65&86.31&76.73&77.08&88.14&76.51&79.23&70.36&71.19&\textbf{11.18M}&\textbf{1.820}\\
    +SE \cite{Jie2019Squeeze} &78.65&90.90&76.58&77.93&87.46&77.26&79.03&69.54&71.33&11.27M&1.821\\
    +SPA \cite{guo2020spanet} &79.17&89.66&\textbf{77.37}&76.91&87.39&76.24&79.13&70.70&71.40&12.14M&1.822\\
    +CA \cite{hou2021coordinate} &77.60&89.62&76.27&76.91&86.49&76.10&79.54&69.07&72.38&11.32M&1.822 \\ \hline
    +SA \cite{woo2018cbam} & 77.60&89.23&75.24&77.42&87.23&75.93&79.34&70.07&71.77&11.18M&1.824 \\
    +NL \cite{wang2018non}&73.96&86.76&71.32&69.10&77.24&64.47&78.02&67.75&69.30&11.97M&1.935\\
    +EA \cite{guo2022beyond}&70.31&86.66&62.73&74.19&83.64&70.78&78.93&64.80&71.55&11.43M&1.915\\
    +GC \cite{cao2019gcnet}&77.08&86.64&74.49&73.51&84.47&70.54&79.03&71.62&71.35&11.36M&1.824 \\ +PPCA&\textbf{80.21}&\textbf{91.81}&77.11&\textbf{78.78}&\textbf{88.56}&\textbf{77.89}&\textbf{80.85}&\textbf{73.52}&\textbf{74.22}&\textbf{11.18M}&\textbf{1.820}\\ \hline \hline
    ResNet50 & 72.92&86.93&69.22&76.57&86.94&69.40&78.33&66.08&69.50&\textbf{23.52M}&4.116 \\
    +SE \cite{Jie2019Squeeze} &73.96&85.64&68.80&77.08&86.73&75.74&77.52&64.40&67.70&26.05M&4.118\\
    +SPA \cite{guo2020spanet}&74.48&84.21&69.84&77.42&87.12&76.71&78.23&67.12&69.27&51.19M&4.153\\
    +CA \cite{hou2021coordinate}&75.52&89.25&73.61&77.76&87.41&77.19&78.73&66.45&70.71&27.33M&4.171 \\ \hline
     +SA \cite{woo2018cbam} &73.96&85.77&70.49&76.57&75.61&86.48&79.03&64.78&70.98&\textbf{23.52M}&4.133\\
    +NL \cite{wang2018non}&65.63&67.22&54.27&76.74&86.91&75.54&77.52&57.81&67.70&46.17M&7.815\\
    +EA \cite{guo2022beyond}&72.92&87.36&70.81&73.68&83.70&70.37&78.83&66.83&72.05&25.46M&4.816 \\
    +GC \cite{cao2019gcnet}&68.23&81.07&58.82&72.67&82.55&69.10&77.92&65.38&68.24&28.58M&4.120 \\
+PPCA&\textbf{78.13}&\textbf{89.27}&\textbf{77.44}&\textbf{79.12}&\textbf{89.14}&\textbf{78.66}&\textbf{79.94}&\textbf{71.71}&\textbf{72.96}&\textbf{23.52M}&4.116\\ 
    \hline
  \end{tabular}

  \label{tab:1}
\end{table*}

\begin{figure*}[htb]
  \centering
    {\includegraphics[width=0.4\linewidth,height=4.5cm]{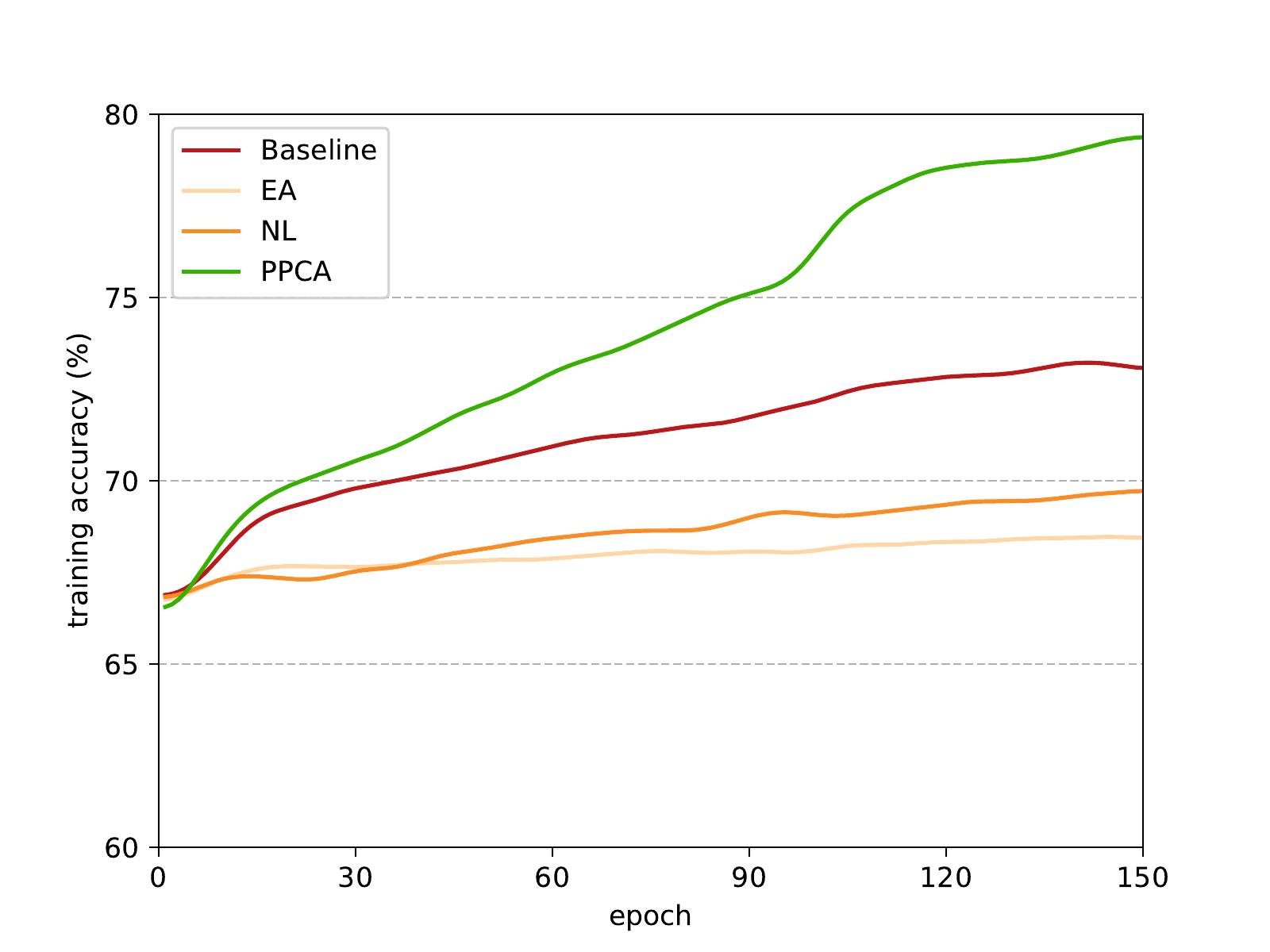}}
    {\includegraphics[width=0.4\linewidth,height=4.5cm]{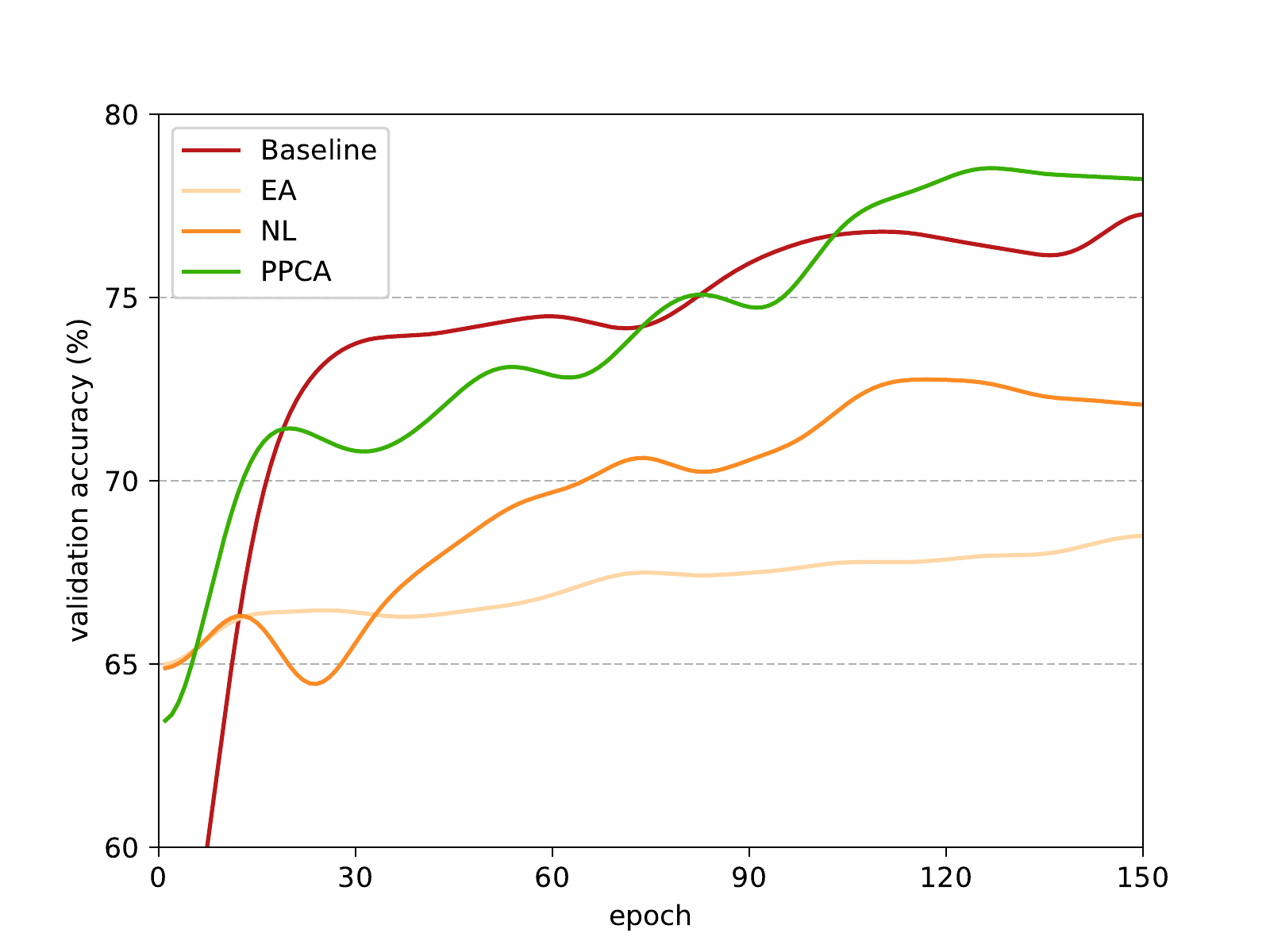}}
  \caption{Training (left) and validation (right) curves on ISIC2018 dataset with ResNet18 (baseline) and different spatial attention methods.}
  \label{fig:2-1}
\end{figure*}

\subsection{Comparisons with advanced attention methods}
\label{sec:class}

\textbf{Performance comparison on ISIC2018 dataset.} 
As shown in Table~\ref{tab:1}(left), PPCA also generally performs better than other SOTA attention methods by balancing performance and model complexity. Remarkably, PPCA outperforms SA (CBAM) by absolute over \textbf{2.60\%} and \textbf{4.10\%} in accuracy under ResNet18 and ResNet50. Compared with comparable SOTA self-attention counterparts (e.g., GC, NL, MHA and EA), PPCA consistently obtains over \textbf{3.1\%} and \textbf{2.6\%} gains of accuracy and F1 accordingly while benefiting fewer computational costs and parameters. For example, NL is \textbf{96\%} larger in parameters and \textbf{90\%} larger in computational cost than PPCA based on ResNet50. We also observe that PPCA outperforms SPA by \textbf{3.65\%} of accuracy, \textbf{5.06\%} of AUC, and \textbf{7.6\%} of F1 accordingly based on ResNet50. This demonstrates the superiority of multi-scale pixel context aggregation via CCPP for spatial attention mechanism compared to multi-scale spatial context aggregation for channel attention mechanism via spatial pyramid pooling. 

Fig.~\ref{fig:2-1} shows the training and validation curves of ResNet18 with PPCA, EA, and NL. During the training process, we can observe that PPCA achieves higher accuracy than baseline, EA, and NL, while EA and NL have lower accuracy than baseline. \textit{This suggests that using multi-scale pixel context information in a pixel-independent way with PPCA is more effective than modeling pixel context long-range dependencies with EA and NL.}

\textbf{Performance comparison on IMED-MD dataset.} Table~\ref{tab:1}(Left) offers the classification results of PPCA and other comparable attention methods based on ResNet18 and ResNet50. It can be observed that our PPCA consistently improves its performance over comparable attention methods. For example, compared to SPA and NL, PPCA achieves absolute over 1.17\% gains of three evaluation measures under ResNet18. The results prove the superiority of PPCA in an adaptive aggregation of multi-scale pixel context information in a pixel-independent manner.


\textbf{Performance comparison on Fundus-Isee dataset.} 
The results also show that PPCA consistently improves performance over SOTA attention methods under fewer budgets, as shown in Table~\ref{tab:2}(Right). PPCA improves two backbones over 2.25\% of F1 and \textbf{3.0\%} of AUC by using almost the same model complexity. In contrast, NL, GC, and EA perform worse than ResNet18, demonstrating that our PPCA is more able to locate significant subtle lesion regions than these self-attention-based spatial attention methods.

\begin{table*}[t]
 \caption{Performance comparison of different attention methods on three MedMNIST datasets (OCTMNIST, RetinaMNIST, and BreastMNIST) in terms of  accuracy, AUC, F1, parameters, and GFLOPs.}
  \centering
  \begin{tabular}{cccc|ccc|ccc|cc}
    \hline
    \multirow{2}{*}{Method}& \multicolumn{3}{c}{OCTMNIST}& \multicolumn{3}{c}{RetinaMNIST} & \multicolumn{3}{c}{BreastMNIST}&\multirow{2}{*}{Params}& \multirow{2}{*}{GFLOPs}\\
     & ACC&AUC&F1&ACC&AUC&F1&ACC&AUC&F1\\
    \hline
    ResNet18 &76.20&93.89&73.61&52.00&68.96&48.52&85.90&87.93&85.39&11.17M&0.458\\
    +SE \cite{Jie2019Squeeze}&77.50&94.89&75.14&51.00&67.62&\textbf{50.95}&85.26&88.01&84.65&11.31M&0.458\\
     +SPA \cite{guo2020spanet}&78.60&94.74&76.91&52.00&73.59&49.54 &82.05&86.77&81.59&12.13M&0.459\\
    +CA \cite{hou2021coordinate}&78.70&94.20&76.27&50.00&\textbf{74.31}&47.65&85.26&87.81&\textbf{86.97}&11.31M&0.459\\ \hline
    +SA \cite{woo2018cbam}&72.70&96.20&68.04&51.50&67.63&50.09&84.62&76.24&84.73&11.17M&0.459\\
    +NL \cite{wang2018non}&75.60&94.11&72.26&51.75&74.19&50.04&82.69&87.98&83.08&11.96M&0.489\\
    +EA \cite{guo2022beyond}&71.60&93.97&65.88&49.25&72.44&43.40&73.72&54.75&63.19 &11.42M&0.482\\
    +GC \cite{cao2019gcnet}&73.20&93.61&68.84&51.50&65.77&48.96&82.05&81.89&86.85&11.35M&0.458\\
    +PPCA &\textbf{79.80}&\textbf{96.33}&\textbf{77.79}&\textbf{53.00}&73.63&50.01&\textbf{87.20}&\textbf{88.89}&\textbf{86.97}&11.17M&0.458\\\hline \hline
    ResNet50 &75.40&92.86&72.04&51.50&69.36&\textbf{50.56}&83.33&88.24&83.30&23.51M&1.053 \\
    +SE \cite{Jie2019Squeeze}&72.50&92.69&67.69&47.75&67.29&47.22&84.62&86.90&84.36&26.24M&1.057\\
    +SPA \cite{guo2020spanet}&77.20&95.04&74.61&50.25& 66.36& 47.63&82.05&88.16&82.59&51.18M&1.085\\
    +CA \cite{hou2021coordinate}&77.60&93.71&74.58&51.00&69.34&50.37&84.62&89.22&84.36&27.32M&1.083 \\ \hline
     +SA \cite{woo2018cbam}&75.50&\textbf{96.36}&71.25&50.50&72.42&48.94&79.49&80.29&75.48&23.51M&1.059\\
    +NL \cite{wang2018non}&65.80&90.40&58.78&48.00&65.18&39.70&76.28&70.38&71.37&46.16M&2.033\\
  +EA~\cite{guo2022beyond}&73.00&91.90&68.36&49.00&70.92&43.84&73.71&63.31&68.89&25.44M&1.223 \\
    
    +GC \cite{cao2019gcnet}&67.70&90.41&60.51&52.75&66.35&49.33&80.77&75.56&77.84 &28.57M&1.060\\
    +PPCA&\textbf{81.90}&95.38&\textbf{79.97}&\textbf{53.25}&\textbf{73.68}&50.51&\textbf{88.46}&\textbf{89.35}&\textbf{84.37}&23.51M&1.053\\ 
    \hline
  \end{tabular}
  \label{tab:2}
\end{table*}

\textbf{Performance comparison on MedMNIST datasets.} According to Table~\ref{tab:2}, PPCA achieves a better trade-off between effectiveness and efficiency than SOTA attention methods on three MedMNIST datasets. Moreover, we observe that the accuracies of all attention methods are low on RetinaMNIST, indicating it is a challenging task as denoted in literature~\cite{yang2021medmnist}.
It is worth noting that PPCA outperforms NL by absolute over \textbf{16\%} on the OCTMNIST dataset by taking ResNet50 as the backbone, although NL is \textbf{96\%} larger in parameters. 

According to Table~\ref{tab:1} and Table~\ref{tab:2}, we note that other competitive attention methods, e.g., SE and SA, do not perform better than baselines except for our PPCA on six medical datasets, verifying the generalization of our method. The following reasons can explain this phenomenon: 1) Dataset scale size of medical images is limited, which leads to the networks easy to be overfitting; 2) other attention methods can not guide the networks to focus on subtle lesion regions well due to context aggregation and context dependency modeling. Overall, The results demonstrate that PPCA effectively leverages the potential of multi-scale pixel context information and pixel normalization to dynamically re-estimate the relative importance of each pixel position in a pixel-independent manner, agreeing with our expectation.

\subsection{Comparisons with different losses}

Table~\ref{tab:4} offers the classification results of CE, SCL, and HL on the ISIC2018 dataset. It can be observed that CE slightly outperforms SCL in accuracy and F1, while SCL performs better than CE in the AUC. Noticeably, HL significantly outperforms individual CE and SCL, demonstrating that HL can take advantage of CE and SCL to boost the performance from both sample-wise and contrastive pair perspectives. Moreover, we also see that setting $\lambda$ value is significant for achieving promising performance.

\begin{table}
  \caption{Performance comparison of different losses on ISIC dataset by taking PPCANet-18 as the backbone}
  \centering
  \begin{tabular}{cccc}
    \hline
    Method & ACC&AUC&F1\\
    \hline
     CE ($\lambda =1$) &80.21&91.81&77.11\\
     SCL ($\lambda =0$) &79.14&92.18&76.48\\
     HL ($\lambda =0.9$) &83.33&95.29&82.28\\
     HL ($\lambda =0.8$) &82.81&95.19&82.00\\
     HL ($\lambda =0.7$) &83.33&95.17&82.49\\
     HL ($\lambda =0.6$) &82.81&\textbf{95.35}&81.97\\
     HL ($\lambda =0.5$) &\textbf{83.85}&95.13&\textbf{82.92}\\
    \hline
  \end{tabular}
  \label{tab:4}
\end{table}

\subsection{Comparisons with SOTA Deep Neural Networks}

Table~\ref{tab:3} offers the classification results of our PPCANet and recent advanced deep neural networks on ISIC2018, iMED-MD, and Fundus-Isee datasets. We observe that our PPCANet generally achieves the best performance among all deep neural networks on these three datasets. For example, our PPCA outperforms transformer and MLP-like architectures like ViT, Swin-T, and MLP-Mixer by 2.61\% in accuracy and 2.41\% in F1 on the ISIC2018 dataset, respectively. We also see that hybrid loss further boosts the medical image classification performance of PPCANet. It improves accuracy, AUC, and F1 of PPCANet by an absolute over 1.8\% on the Fundus-Isee dataset.


Table~\ref{tab:3-1} presents the results of PCANet and other advanced deep neural networks on three MedMNIST datasets. Due to the input image size problem, some SOTA deep neural networks in Table~\ \ref{tab:3-1} are not applicable to MedMNIST datasets. From Table~\ref{tab:3-1}, we can see that our PCANet generally performs better than comparable deep neural networks, and hybrid loss further boosts its performance, keeping consistent with our expectations. For example, PCANet with hybrid loss outperforms RCRNet, CABNet, and PSANet by over \textbf{7\%} of accuracy and F1 on the OCTMNIST dataset. All in all, the results in Table~\ref{tab:3} and Table~\ref{tab:3-1} demonstrate the superiority and generalization of our PPCANet with hybrid loss.

\begin{table*}
  \caption{Performance comparison of our PPCANet and state-of-the-art deep neural networks on two medical image datasets (ISIC2018, iMED-MD, and Fundus-Isee).}
  \centering
  \begin{tabular}{c|ccc|ccc|ccc}
    \hline
    \multirow{2}{*}{Method}& \multicolumn{3}{c}{ISIC2018}&  \multicolumn{3}{c}{IMED-MD}&\multicolumn{3}{c}{Fundus-Isee}\\
     & ACC&AUC&F1&ACC&AUC&F1&ACC&AUC&F1\\
    \hline
     VGGNet16&79.69&88.79&77.39&78.10&88.19&77.68&80.24&76.60&73.26\\
     ConvNeXt~\cite{liu2022convnet} &64.06&64.95&50.03&76.91&87.89&75.68&77.52&56.19&67.70\\
    ResNet18 &78.65&86.31&76.73&77.08&88.14&76.51&79.23&70.36&71.19\\
    ResNet50&72.92&86.93&69.22&76.57&86.94&69.40&78.33&66.08&69.50\\\hline
    CABNet~\cite{he2020cabnet} &78.65&85.79&78.04&77.08&87.17&75.72&79.84&71.31&72.78\\
    RIRNet ~\cite{ZHANG2022102499}&78.65&89.69&77.34&73.84&83.12&71.02&79.74&70.65&72.35\\
    RCRNet~\cite{zhang2024regional}&72.40&81.59&69.14&72.84&83.63&68.90&77.52&62.68&67.70\\
    Lian et al. \cite{9369104} &77.60&88.37&76.00&77.59&88.37&76.63&79.03&69.13&71.33\\
    DANet~\cite{fu2020scene}&73.44&85.99&68.32&76.74&87.13&75.85&78.63&68.35&70.11\\
    MANs~\cite{9378801}&78.65&90.58&76.24&78.27&87.43&77.57&80.04&72.29&72.69 \\
    PSANet~\cite{9229188}&79.17&82.67&75.99&73.51&84.86&71.54&79.64&72.34&72.32\\
     CoAtNet~\cite{dai2021coatnet} &77.08&88.89&74.94&73.01&84.52&68.70&79.84&70.07&72.49\\
    AANet~\cite{Bello_2019_ICCV}  &70.31&87.90&65.39&64.35&73.86&59.60&77.52&49.49&67.70\\
    ViT~\cite{dosovitskiy2020image} &77.60&91.41&74.07&73.68&81.77&70.19&78.73&67.93&71.82\\
    SSFormer~\cite{wang2022stepwise}&73.96&87.10&70.93&76.40&86.97&75.60&77.52&50.05&67.70\\
    Swin-T~\cite{liu2021swin}  &77.60&90.75&74.70&77.76&86.31&76.72&78.83&64.73&70.93\\
   PVTv2~\cite{wang2021pvtv2}  &73.96&89.95&70.41&75.21&85.79&74.25&77.52&51.67&67.70\\ 
   MetaFormer~\cite{yu2022metaformer}&79.17&91.78&76.03&74.87&84.17&71.30&80.65&74.04&73.94\\  

     Res-MLP~\cite{touvron2021resmlp} &68.75&82.06&64.01&75.04&85.67&72.30&77.52&51.06&67.70\\
    MLP-Mixer \cite{tolstikhin2021mlp} &75.52&90.41&73.01&75.55&85.14&73.51&78.73&68.40&70.38\\ \hline 
    PPCANet  &\textbf{80.21}&\textbf{91.81}&\textbf{77.11}&\textbf{78.78}&\textbf{88.56}&\textbf{77.89}&\textbf{80.85}&\textbf{73.52}&\textbf{74.22}\\
    PPCANet+proposed loss  &\textbf{83.85}&\textbf{95.13}&\textbf{82.92}&\textbf{80.14}&87.46&\textbf{79.33}&\textbf{82.66}&\textbf{86.60}&\textbf{77.79}\\
    \hline
  \end{tabular}

  \label{tab:3}
\end{table*}

\begin{table*}
  \caption{Performance comparison of our PPCANet and state-of-the-art deep neural networks on three MedMNIST datasets.}
  \centering
  \begin{tabular}{c|ccc|ccc|ccc}
    \hline
    \multirow{2}{*}{Method}& \multicolumn{3}{c}{OCTMNIST}& \multicolumn{3}{c}{RetinaMNIST} & \multicolumn{3}{c}{BreastMNIST}\\
     & ACC&AUC&F1&ACC&AUC&F1&ACC&AUC&F1\\
    \hline
     VGGNet16&79.60&95.82&77.49&51.13&69.12&49.23&87.18&\textbf{90.58}&83.46\\
     ConvNeXt~\cite{liu2022convnet} &72.60&93.81&67.83&53.20&73.59&50.13&73.08&52.39&61.71\\
    ResNet18 &76.20&93.89&73.61&52.00&68.96&48.52&85.90&87.93&85.39\\
 ResNet50&75.40&92.86&72.04&51.50&69.36&\textbf{50.56}&83.33&88.24&83.30\\\hline
    CABNet~\cite{he2020cabnet} &78.70&94.64&76.93&52.00&70.37&44.78&81.25&83.35&80.45\\
    RIRNet ~\cite{ZHANG2022102499}&76.70&95.76&74.26&51.50&72.71&43.98&86.54&82.77&81.12\\
    RCRNet~\cite{zhang2024regional}&75.70&95.14&73.19&51.75&74.01&48.64&80.77&82.38&79.40 \\
    Lian et al. \cite{9369104} &77.70&94.59&75.37&52.00&68.36&49.99&82.05&85.77&81.22\\
    DANet~\cite{fu2020scene}&74.30&93.87&72.10&52.50&74.16&47.30&75.64&60.32&68.17\\
    MANs~\cite{9378801}&75.30&94.21&72.36&52.00&70.35&44.46&82.05&80.08&81.02\\
    PSANet~~\cite{9229188}&76.10&94.27&72.60&51.25 &66.67&43.78&83.97&88.47&78.45\\
    ViT~\cite{dosovitskiy2020image} &71.40&93.11&66.16&53.00&73.67&49.52&73.08&63.85&61.71\\
   PVTv2~\cite{wang2021pvtv2}&79.50&95.44&77.77&52.50&73.76&50.16&80.13&76.26&77.29\\ \hline

    PPCANet &\textbf{81.90}&95.38&\textbf{79.97}&\textbf{53.25}&73.68&50.51&\textbf{88.46}&89.35&\textbf{84.37}\\
    PPCANet+proposed loss  &\textbf{83.80}&\textbf{97.03}&\textbf{82.76}&\textbf{54.75}&\textbf{75.63}&50.35&\textbf{89.74}&89.41&\textbf{86.58}\\
    \hline
  \end{tabular}

  \label{tab:3-1}
\end{table*}

\begin{figure*}
   
    \centering	\centerline{\includegraphics[width=0.95\linewidth,height=7cm]{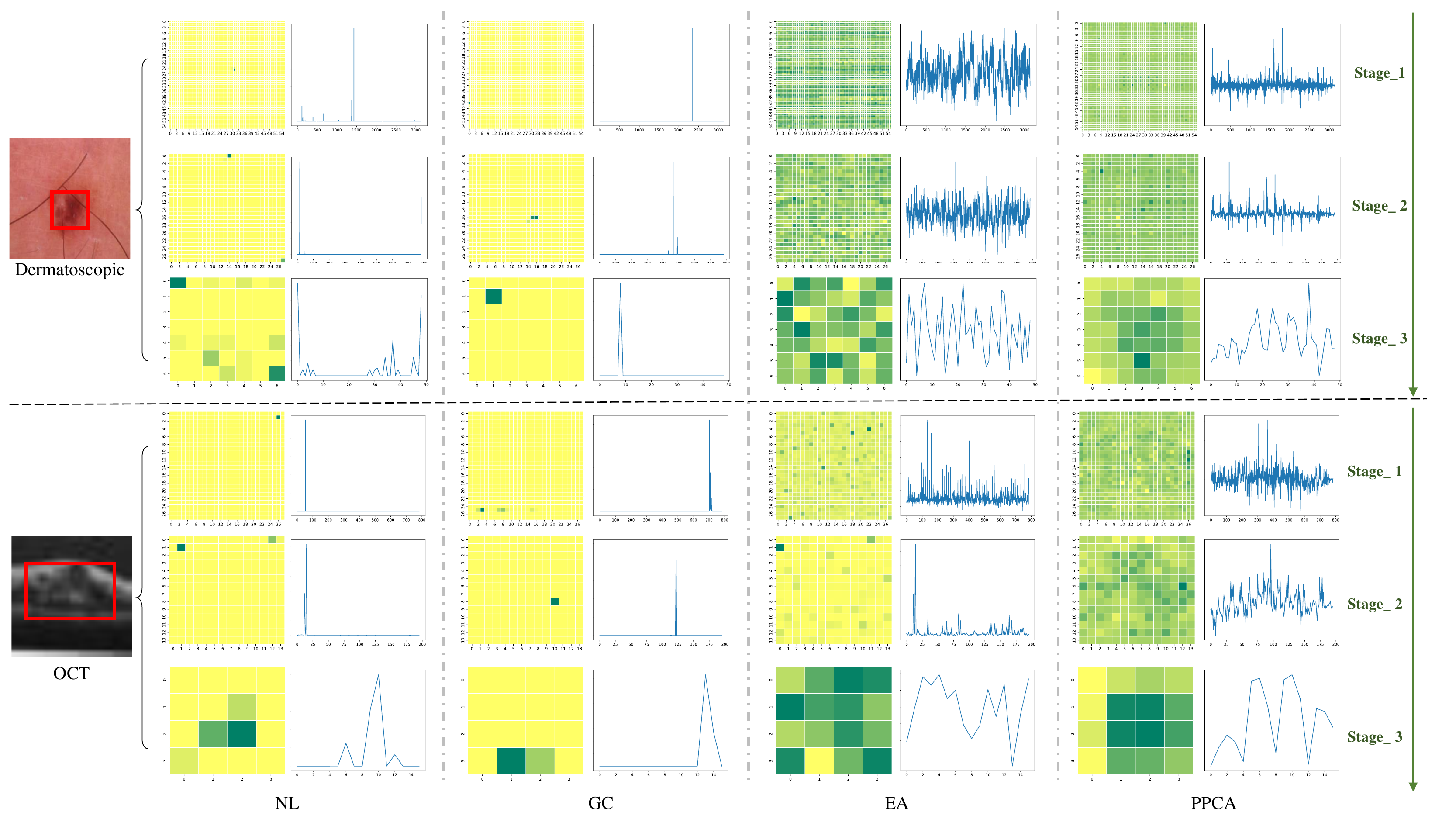}}
	
  \caption{The pixel attention weight feature maps and pixel attention weight distributions of NL, GC, EA, and PPCA at three stages of ResNet18 on ISIC2018 (a) and OCTMNIST (b) datasets.}
  \label{fig:5}
\end{figure*}


\subsection{Visual Analysis and Explanation}

\subsubsection{Attention Weight Visualization} 

Fig.~\ref{fig:5} plots pixel attention weight feature maps and pixel attention weight distributions of NL, GC, EA, and PPCA at three stages based on the ISIC2018 and OCTMNIST datasets: low-level (Stage\_1), middle-level (Stage\_2), and high-level (Stage\_3). Here, we take ResNet18 as the backbone, and red boxes in the dermatoscopic image and OCT image refer to lesion regions. In the pixel attention weight feature maps, the darker the cyan color, the higher the pixel attention weight value, and vice versa. Pixel attention weight distributions indicate corresponding pixel attention weight values for each pixel position from the pixel position perspective, aiming to help the audience understand the. We find that compared with attention weight differences of NL, GC, and EA between non-lesion and lesion regions, the attention weight differences among pixel positions of PPCA are more apparent. That is, our PPCA is more capable of guiding the DNNs to learn subtle lesion region information than other comparable self-attention-based methods. Interestingly, this phenomenon becomes more evident when DNNs go deeper according to pixel attention weight distributions and pixel attention weight feature maps.


We argue the advantages of our PPCA over NL, GC, and EA as three-fold. 
\begin{itemize}
    \item Existing self-attention-based spatial attention methods only use single-scale pixel context information, ignoring the relative importance of multi-scale pixel context information. Hence, we are the first to exploit the potential of multi-scale pixel context information at each pixel position to improve the representational power of CNNs through cross-channel pyramid pooling.

    \item We propose a pixel normalization to eliminate the inconsistency among multi-scale pixel context information, which has not been studied before. 

    \item Existing self-attention-based spatial attention methods capture long-range dependency across all pixel positions for highlighting significant pixel positions, leading to redundant pixel position information introduction. However, our PPCA applies a pixel-independent manner to emphasize significant pixel positions and suppress trivial ones.

\end{itemize}

\subsubsection{Multi-Scale Pixel Context Value Visualization} 
\begin{figure*}
	\centering
	\centerline{\includegraphics[width=0.85\linewidth,height=14cm]{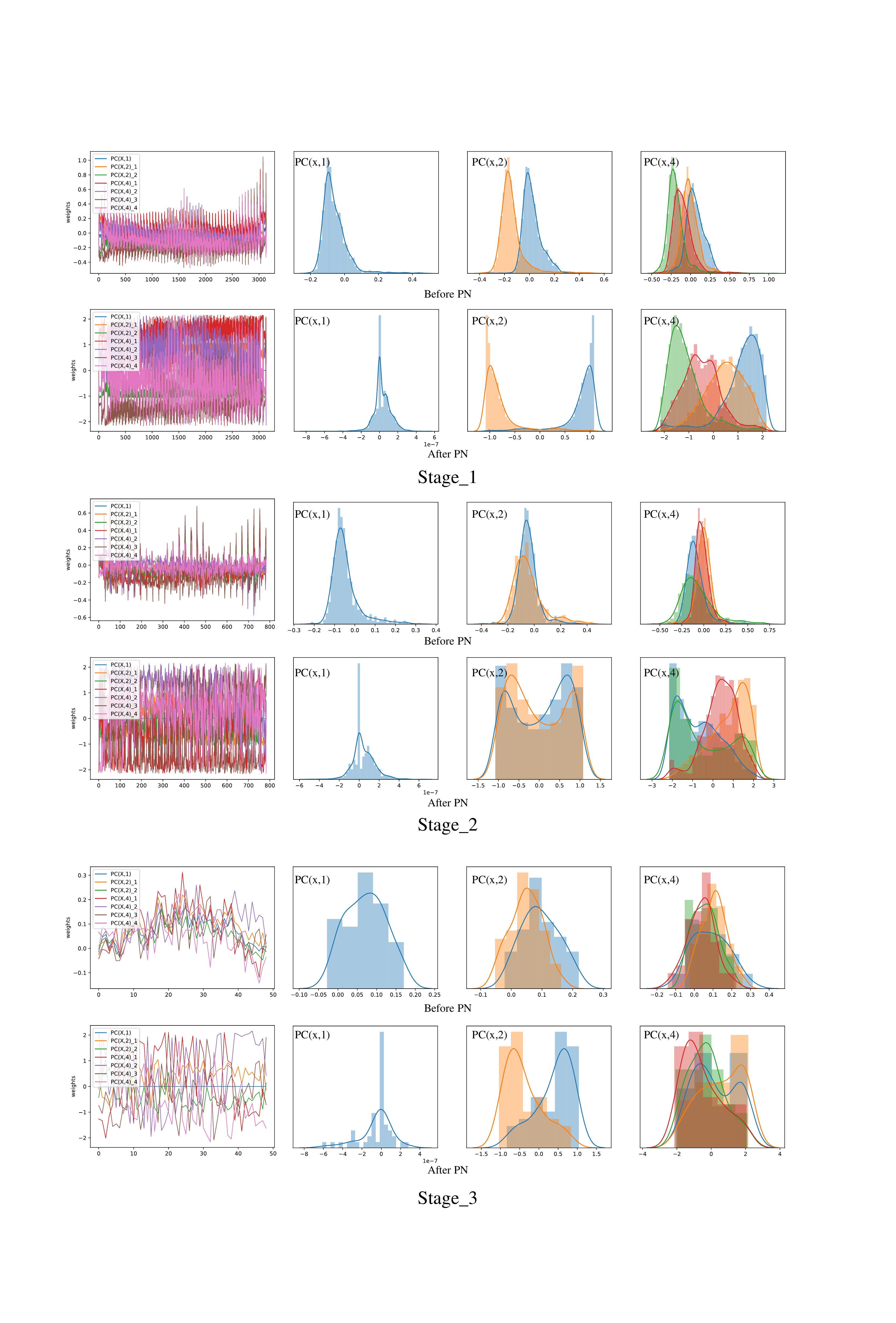}}
	\caption{The multi-scale pixel context feature distributions before and after PN of PPCA based on ResNet18 at three different stages. The dataset is ISIC2018.}
	\label{fig:7}
\end{figure*}

\begin{figure*}
	\centering
	\centerline{\includegraphics[width=0.84\linewidth,height=14cm]{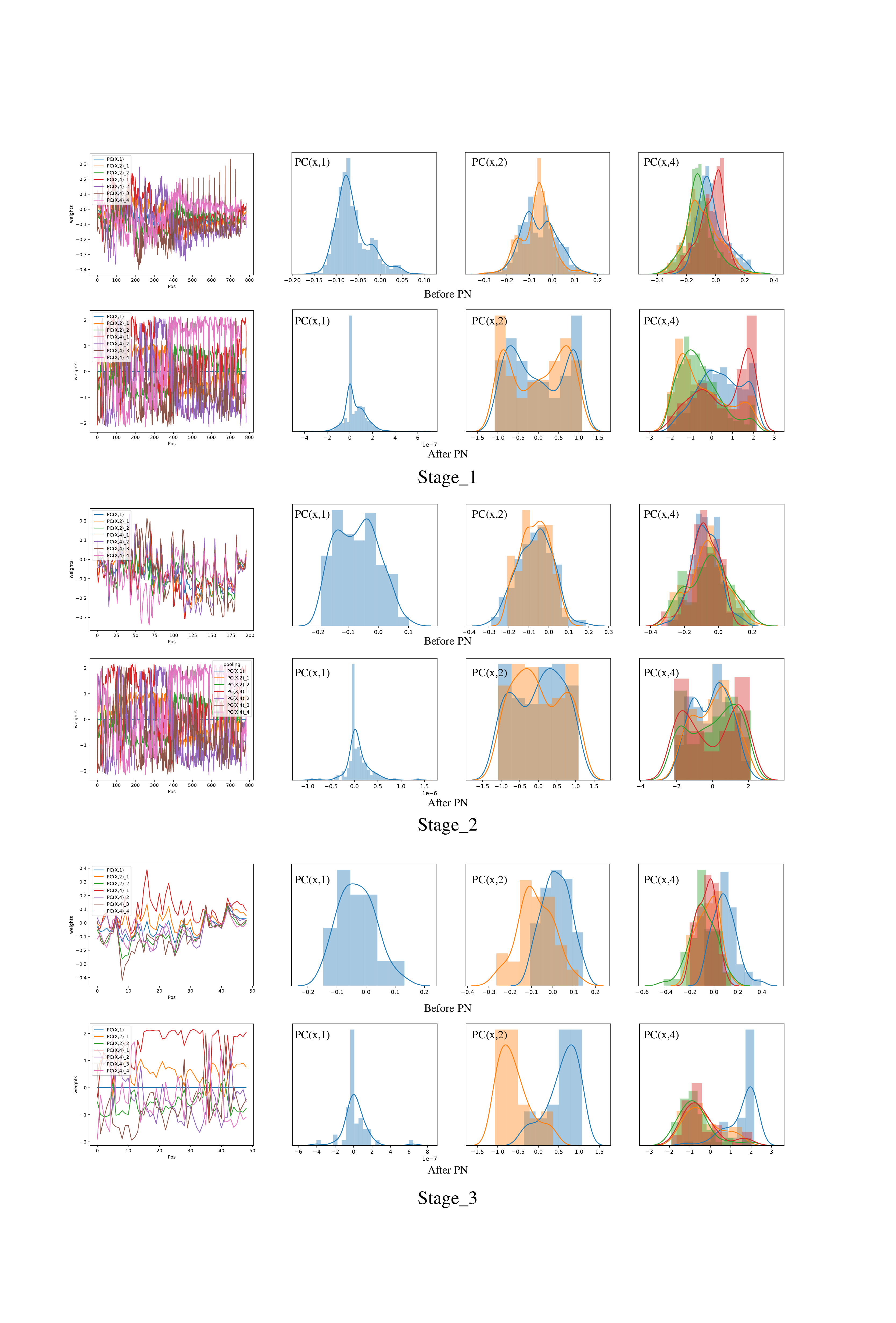}}
	\caption{The pixel attention weight feature maps and pixel attention weight distributions of PPCA based on ResNet18 at three different stages. The dataset is OCTMNIST.}
	\label{fig:8}
\end{figure*}

Fig.~\ref{fig:7} and Fig.~\ref{fig:8} present the multi-scale pixel context feature distributions before and after PN at different stages of PPCA along the pixel position at ISIC2018 and OCTMNIST datasets accordingly. We observe that multi-scale pixel context feature distributions are different from each other, indicating that they play varying significance in PPCA. The fluctuations of multi-scale pixel context feature distribution after PN are smaller than before PN, proving that PN can effectively address the inconsistency among multi-scale pixel context features and demonstrate its generalization applicability on different datasets.

\subsubsection{Multi-Scale Pixel Context Weight Visualization} 

\begin{figure*}[htbp]
	\centering
	\includegraphics[width=0.45\textwidth,height=7.5cm]{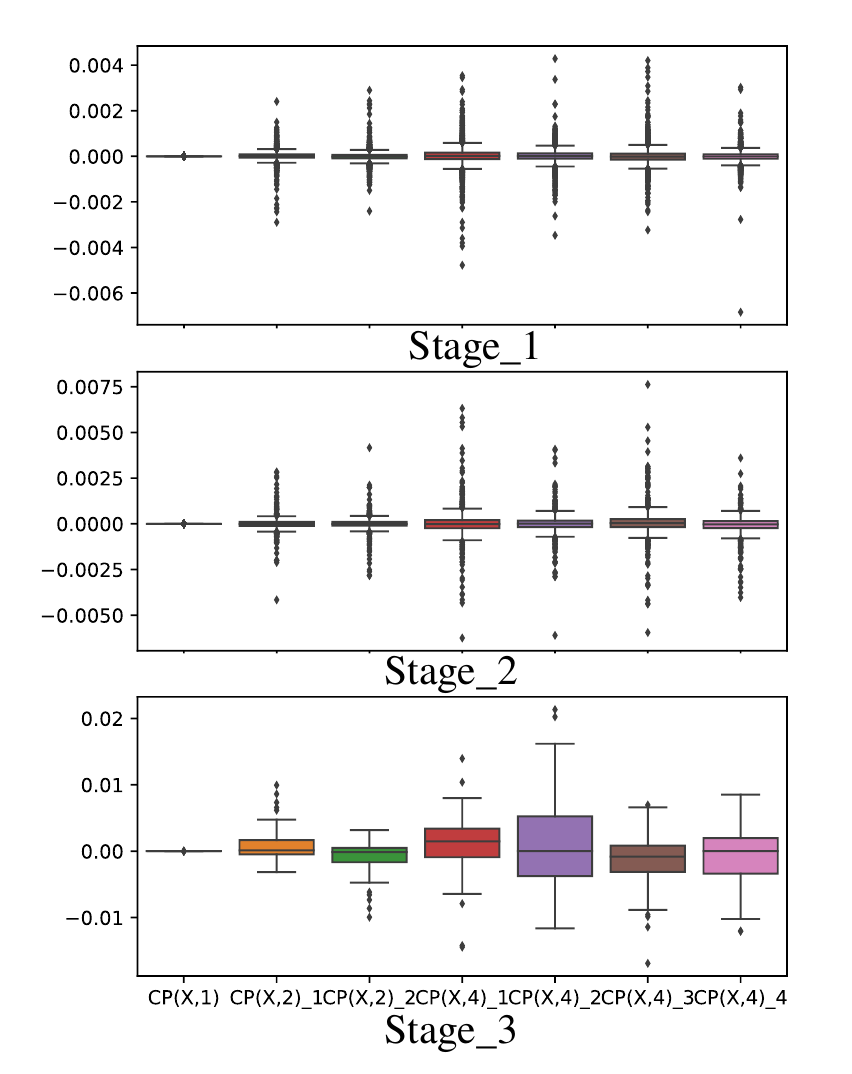} 
	\includegraphics[width=0.45\textwidth,height=7.5cm]{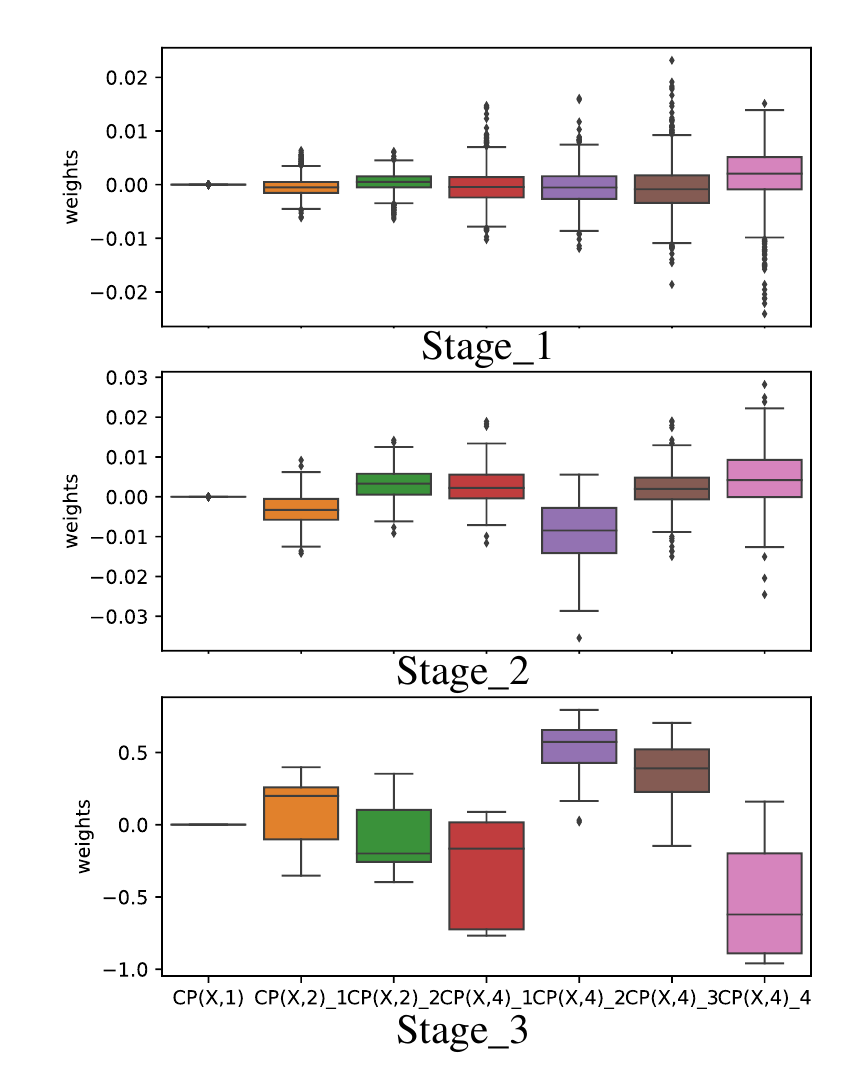} 
        
	\caption{Multi-scale pixel weight distributions of PPCA of ResNet18 at three stages on ISIC2018 and OCTMNIST datasets}
	\label{fig:9}
\end{figure*}

Fig.~\ref{fig:9} offers multi-scale pixel context weight distributions of PPCA based on ResNet18 at three stages. We find a significant difference between multi-scale pixel context weight distributions, proving our PPCA adaptively sets relative weights to multi-scale pixel contexts, guiding deep neural networks to emphasize or suppress significant pixel positions.

\section{Discussion}
\label{sec:5}

\subsection{Ablation Study}
\label{sec:abla}
In this section, we perform a series of ablation experiments to examine which factors affect the effectiveness of PPCA. Throughout the ablation study, we adopt ResNet18 as a backbone architecture and recognize skin disease on ISIC2018 dataset by following the same experiment setting in Section~\ref{sec:3}.

\subsubsection{Effects of Number of Cross-Channel Scales} 
Table~\ref{tab:5} provides results of seven different cross-channel scales used in CCPP for our method. 1 denotes one cross-channel scale: 1$\times$H$\times$W, 2 denotes two cross-channel scales: 1$\times$H$\times$W, and 2$\times$H$\times$W, and so on. PPCA is sensitive to the variation of cross-channel scale number, and this is mainly because mean and variance change with the number of multi-scale pixel context features per pixel position, indicating they play different roles. We find it is challenging to set proper cross-channel scale number. When cross-channel scale=3, it performs better than other settings, which is adopted in this paper.

\begin{table}
  \centering
  \caption{Results of different cross-channel scales with PPCA on the ISIC2018 testing dataset.}
  \begin{tabular}{ccc}
   \hline
    Cross-channel scale & ACC&F1 \\
    \hline
     1 & 78.13 & 75.95\\
     2 & 61.46 & 49.72\\
     3 & \textbf{80.21} & \textbf{77.11}\\
     4 & 78.13 & 76.76\\
     5 & 77.60 & 76.15\\
     6 & 77.60 & 75.55\\
     7 & 77.60 & 76.16\\
    \hline
  \end{tabular}
  
  \label{tab:5}
\end{table}

\subsubsection{Effects of Normalization Methods}
To investigate the effects of the normalization methods on the PPCA module, this paper compares PN with BN, IN, LN, and the original (without any normalization operation). According to Table~\ref{tab:6}, we see that the performance of PPCA with PN is better than PPCA with the other three normalization methods and the original, demonstrating the advantages of PN in eliminating the inconsistency of pixel context feature distribution at the same pixel positions, keep consistent with Fig.~\ref{fig:7} and Fig.~\ref{fig:8}.

\begin{table}
 \caption{Results of different normalization methods with PPCA on ISIC2018 testing dataset}
  \centering
  \begin{tabular}{ccc}
    \hline
    Normalization & ACC&F1  \\
    \hline
    Original & 78.65 & 76.53\\
    BN & 77.60 & 75.62\\
    IN & 78.65 & 77.14\\
    LN & 77.60 & 74.46 \\
    PN (ours) & \textbf{80.21} & \textbf{77.11}\\
    \hline
  \end{tabular}
  \label{tab:6}
\end{table}

\subsubsection{Effects of Initialization Methods} 
Table~\ref{tab:7} lists the results of initialization methods for learnable parameters $W$. We find it more appropriate to initially set $W$ to 0 rather than 1, referring to the classification results. This initialization also conduces to PPCA biasedly to give the higher pixel attention weights for informative pixel positions, as shown in Fig.~\ref{fig:1} and Fig.~\ref{fig:5}.

\begin{table}
  \centering
   \caption{Results of two initialization methods with PPCA on ISIC2018 testing dataset}
  \begin{tabular}{ccc}
    \hline
    \multicolumn{3}{c}{Initialization} \\
    \hline
    W & ACC &F1 \\
    \hline
    0& \textbf{80.21}& \textbf{77.11}\\
    1& 79.69& 77.04\\
    \hline
  \end{tabular} 
  \label{tab:7}
\end{table}

\subsubsection{Effects of Pixel Context Adaption Implementations}

Table~\ref{tab:8} lists the classification results of four different pixel context adaption implementations. We see that PFC outperforms Conv$1\times1$, Conv$5\times5$, and summation operation, verifying the effectiveness of PFC in adjusting the significance of multi-scale pixel context features independently.

\begin{table}
  \centering
   \caption{Results of four pixel context adaption implementations with PPCA on ISIC2018 testing dataset}
  \begin{tabular}{ccc}
    \hline
    \multicolumn{3}{c}{Implementation} \\
    \hline
    & ACC &F1 \\
    \hline
    Conv$1\times1$&79.69&\textbf{77.54} \\
    Conv$5 \times5$&78.65&76.12 \\
    Summation&78.13&76.58 \\
     PFC&\textbf{80.21}& 77.11 \\
    \hline
  \end{tabular} 
  \label{tab:8}
\end{table}

\subsection{Validation}

\textbf{Natural Image Classification on CIFAR Benchmarks.}  CIFAR benchmarks contain CIFAR-10 and CIFAR-100 \cite{krizhevsky2009learning}, which are colored natural images of 32$\times$32 pixels. Training and testing datasets provide 50,000 and 10,000 images accordingly. We follow the standard practice to augment image data by padding zero to four pixels and randomly cropping them to the original size. As listed in Table~\ref{tab:9}, PPCA significantly improves the performance on CIFAR benchmarks with minimal parameter and computational cost increment, which proves the generalization ability of PPCA is not constrained to medical image datasets.

\begin{table}
\begin{center}
 \caption{Performance comparison of different attention methods on CIFAR benchmarks in terms of accuracy, parameters, and GFLOPs.}
\label{tab:9}
\begin{tabular}{c|c|c|c|c}
\hline
  \multirow{2}{*}{Method} & CIFAR-10& CIFAR-100&\multirow{2}{*}{Params}& \multirow{2}{*}{GFLOPs}\\ 
  &ACC &ACC \\ \hline
  ResNet18&93.02 &74.56&\textbf{11.22M}&\textbf{0.557}\  \\
  +SE \cite{Jie2019Squeeze}&94.84 &75.19&11.32M&0.557 \\
  +SPA \cite{guo2020spanet}&95.00 &75.56&12.18M&0.557 \\
   +CA \cite{hou2021coordinate}&95.21&77.73&11.36M&0.558\\
  +NL \cite{wang2018non}&93.38&71.97&12.01M&0.595\\
  +GC \cite{cao2019gcnet}&95.38&77.53&11.40M&0.557\\
  +EA \cite{guo2022beyond}&93.16&72.05&11.47M&0.588\\
  +PPCA &\textbf{95.56}&\textbf{78.70}&\textbf{11.22M}&\textbf{0.557}\\
  \hline
   ResNet50&93.62&78.51&\textbf{23.71M}&\textbf{1.305} \\
  +SE \cite{Jie2019Squeeze}&95.35 &79.28&26.24M &1.309 \\
   +SPA \cite{guo2020spanet}&94.63&78.21&51.37M &1.338 \\
  
  +CA \cite{hou2021coordinate}&95.52&79.45&27.51M&1.340\\
  +NL \cite{wang2018non}&94.00&72.15&46.36M&2.515\\
  +GC \cite{cao2019gcnet}&95.60&78.37&28.77M&1.312\\
  +EA \cite{guo2022beyond}&93.98&71.85&25.64M&1.536\\
  +PPCA &\textbf{95.92}&\textbf{79.93}&\textbf{23.71M}&\textbf{1.305}\\
 \hline
\end{tabular}
\end{center}
\end{table}

\textbf{Object Detection on COCO.} We investigate the effectiveness of PPCA on the object detection task with the COCO 2017 \cite{lin2014microsoft} dataset (train set with 118k images and validation set with 5k images). We follow the commonly used object detection settings of Faster R-CNN by taking ResNet50 as the baseline and adopting commonly used evaluation measures. As shown in Table~\ref{tab:10}, our PPCA achieves better object detection results than other SOTA methods, verifying its effectiveness in the object detection task.

\begin{table}
\begin{center}
 \caption{Objection detection results of different  methods on COCO benchmarks by using Faster R-CNN as detector.}
\label{tab:10}
\begin{tabular}{c|c|c|c|c|c|c}
\hline
Method & AP & $AP_{50}$&$AP_{75}$& $AP_{S}$& $AP_{M}$&$AP_{L}$\\ \hline
  ResNet50& 36.4& 58.2& 39.2& 21.8& 40.0& 46.2\\
  +SE &37.7&60.1&40.9&\textbf{22.9}&41.9&48.2\\
  +NL & 37.4& 59.1& 40.4&21.7 & 41.1&\textbf{49.4}\\
   +PPCA & \textbf{38.2}& \textbf{60.5}& \textbf{41.3}& 22.6& \textbf{42.2}&49.0\\
 \hline
\end{tabular}
\end{center}
\end{table}

\textbf{Medical Segmentation on Synapse.}
Finally, we provide the medical image segmentation results of our proposed PPCA by using UNet on the Synapse dataset  \cite{landman2015miccai}. It is comprised of 30 CT scans for the multi-organ segmentation task. In this paper, we follow the same dataset splitting strategies and the evaluation metrics used in \cite{10183842}: Dice and average $95\%$ Hausdorff distance (HD95). As shown in Table~\ref{tab:11}, our PPCA significantly performs better than existing SOTA segmentation methods. These results demonstrate the generalization ability of our method in other learning tasks.

\begin{table}
\begin{center}
 \caption{Performance comparisons of previous SOTA segmentation methods and our PPCA on multi-organ segmentation task.}
\label{tab:11}
\begin{tabular}{c|c|c}
\hline
Method & Dice & HD95 \\ \hline
  UNet~\cite{he2023h2former}& 67.89& 26.60\\
  UNet++~\cite{he2023h2former}&68.50&42.39\\
  Att-UNet\cite{he2023h2former}&67.40&35.73\\
  PSPNet\cite{he2023h2former}&67.74&30.28\\
  SSFormer~\cite{wang2022stepwise}&62.88&21.34\\
  mmFormer~\cite{10183842}&69.76 &20.55\\
  DeepLabv3+&66.53&29.58\\
  Ours &\textbf{72.62}&\textbf{18.48}\\
 \hline
\end{tabular}
\end{center}
\end{table}

\subsection{Limitations and Future Work}

This paper argues that the spatial attention mechanisms, especially self-attention-based methods, often achieve promising performance in natural image-based tasks but may not perform well in medical image analysis. In seeking answers to this phenomenon, we find that long-range dependency capturing and pixel context aggregation have significant effects, which most existing works have ignored. To tackle these two questions, we propose an alternative method to highlight informative pixel positions and suppress trivial ones without capturing long-range dependency among pixel positions and then design how to embed the multi-scale pixel context features into the spatial attention module. Particularly, we propose a PPCANet for automatic medical image classification by exploiting the potential of multi-scale pixel context information in a pixel-independent manner. Although our method performs better than recent SOTA methods on several learning tasks and provides a visual explanation of DNN in the decision process, this paper still has some limitations, which are concluded as follows:

\begin{itemize}
    \item This paper only adopts the averaged cross-channel pyramid pooling (CCPP) method to obtain multi-scale pixel context features, and other CCPP methods can be designed to extract other multi-scale pixel context feature types for further improving performance.
    \item We propose the pixel normalization (PN) method to eliminate the fluctuation of multi-scale pixel context distribution per pixel position, which still has improvement room. For example, we can combine PN with other normalization methods, e.g., BN and GN, to further improve the performance of PPCANet. Moreover, the theoretical basis of PN is insufficient, we plan to tackle this limitation in the improved PN method.
    \item We only test the effectiveness of PPCANet on 2D medical image classification tasks and 2D medical image segmentation task due to the constrained computing resource.
\end{itemize}

To address the above limitations, we plan to improve the architecture design of PPCANet by prompting the idea of our PPCA module. Furthermore, we will apply 3D medical image analysis and other computer vision tasks to test the generalization ability and effectiveness of our method.

\section{Conclusion}
\label{sec:6}

In this paper, we propose an efficient yet lightweight pyramid pixel context adaption module (PPCA) to dynamically estimate the relative significance of each pixel position in a pixel-independent manner based on aggregated multi-scale pixel context features. By incorporating multi-scale pixel context features into feature maps at the pixel-independent level, it improves the representational ability of a CNN efficiently. 
Furthermore, we also utilize supervised contrastive loss combined with CE  to exploit the label information for improving performance from both contrastive pair and sample-wise aspects. The comprehensive results on six medical image classification datasets, CIFAR datasets, COCO 2017 dataset, and Synapse dataset demonstrate the effectiveness and generalization ability of our PPCA through comparisons to SOTA methods. Furthermore, we provide visual analyses and ablation studies to explain the significance of PPCA in adjusting the relative contributions of multi-scale pixel context information and pixel normalization, conducing to improving the interpretability of CNNs. In future work, we plan to develop more efficient methods to explore multi-scale pixel context information, which may provide new insights into spatial attention design. We hope our efficient and lightweight design sheds light on future research on attention methods.


\bibliographystyle{IEEEtran}
\bibliography{ref.bib}



\end{document}